\documentclass[acmtog,nonacm]{acmart}

\usepackage{multirow}
\usepackage{pifont}
\newcommand{\cmark}{\ding{51}}

\begin{document}

\setcopyright{none}
\acmJournal{TOG}
\acmYear{2026} \acmVolume{45} \acmNumber{6} \acmArticle{178}
\acmMonth{12} \acmDOI{10.1145/3842577}

\title{PartLLM: A Unified Multimodal Foundation for 3D Part Segmentation}

\author{Zhe Zhu}
\affiliation{
  \institution{Taiyuan University of Technology, Tencent Visvise}
  \country{China}}
\email{zhuzhe0619@gmail.com}

\author{Yiheng Zhang}
\affiliation{
  \institution{Hong Kong University of Science and Technology, Tencent Visvise}
  \country{China}}
\email{e1349382@u.nus.edu}

\author{Peng Li}
\affiliation{
  \institution{Taiyuan University of Technology}
  \country{China}}
\email{pengli0206@gmail.com}

\author{Zixing Zhao}
\affiliation{
  \institution{Tencent Visvise}
  \country{China}}
\email{zixingzhao@tencent.com}

\author{Honghua Chen}
\affiliation{
  \institution{Lingnan University}
  \country{China}}
\email{honghuachen@LN.edu.hk}

\author{Yaqing Zhang}
\affiliation{
  \institution{Tencent Visvise}
  \country{China}}
\email{yaqingzhang@tencent.com}

\author{Le Wan}
\affiliation{
  \institution{Tencent Visvise}
  \country{China}}
\email{vinowan@tencent.com}

\author{Zhiyang Dou}
\affiliation{
  \institution{MIT}
  \country{USA}}
\email{frankdou@mit.edu}

\author{Cheng Lin}
\authornotemark[3]
\affiliation{
  \institution{Macau University of Science and Technology}
  \country{China}}
\email{chlin@connect.hku.hk}

\author{Yuan Liu}
\authornotemark[2]
\affiliation{
  \institution{Hong Kong University of Science and Technology}
  \country{China}}
\email{liuyuanwhuer@gmail.com}

\author{Mingqiang Wei}
\authornotemark[2]
\affiliation{
  \institution{Taiyuan University of Technology}
  \country{China}}
\email{mingqiang.wei@gmail.com}

\author{Wenping Wang}
\affiliation{
  \institution{Texas A\&M University}
  \country{USA}}
\email{wenping@tamu.edu}

\thanks{\textsuperscript{\textdaggerdbl} Project lead.\quad
  \textsuperscript{\textdagger} Corresponding authors.}

\begin{abstract}
Part segmentation is a fundamental problem in computer graphics and 3D vision.
Recent works have expanded 3D part segmentation beyond fixed
taxonomies, but existing approaches typically only address
a specific setting, such as text-guided part segmentation or point-based
interaction.
In this work, we argue that these settings can be unified as an intent-conditioned generative problem, where different prompts specify the desired part decomposition. 
To this end, we introduce PartLLM, a unified multimodal model that formulates 3D part segmentation as autoregressive semantic decomposition.
Conditioned on an input shape and a user prompt, PartLLM autoregressively generates semantic part hypotheses as queries for mask prediction and feeds them to a decomposition-aware decoder that jointly predicts coherent part masks.
This unified design supports text-guided part segmentation, interactive
segmentation, and full-shape semantic decomposition with controllable
granularity within a single model.
Extensive experiments across these task settings show that PartLLM consistently outperforms task-specific baselines, demonstrating the effectiveness of unifying 3D part segmentation under an intent-conditioned generative formulation.
Project Page: \url{https://czvvd.github.io/PartLLMPage/}.

\begin{figure*}[t]
  \centering
  \includegraphics[width=0.985\textwidth]{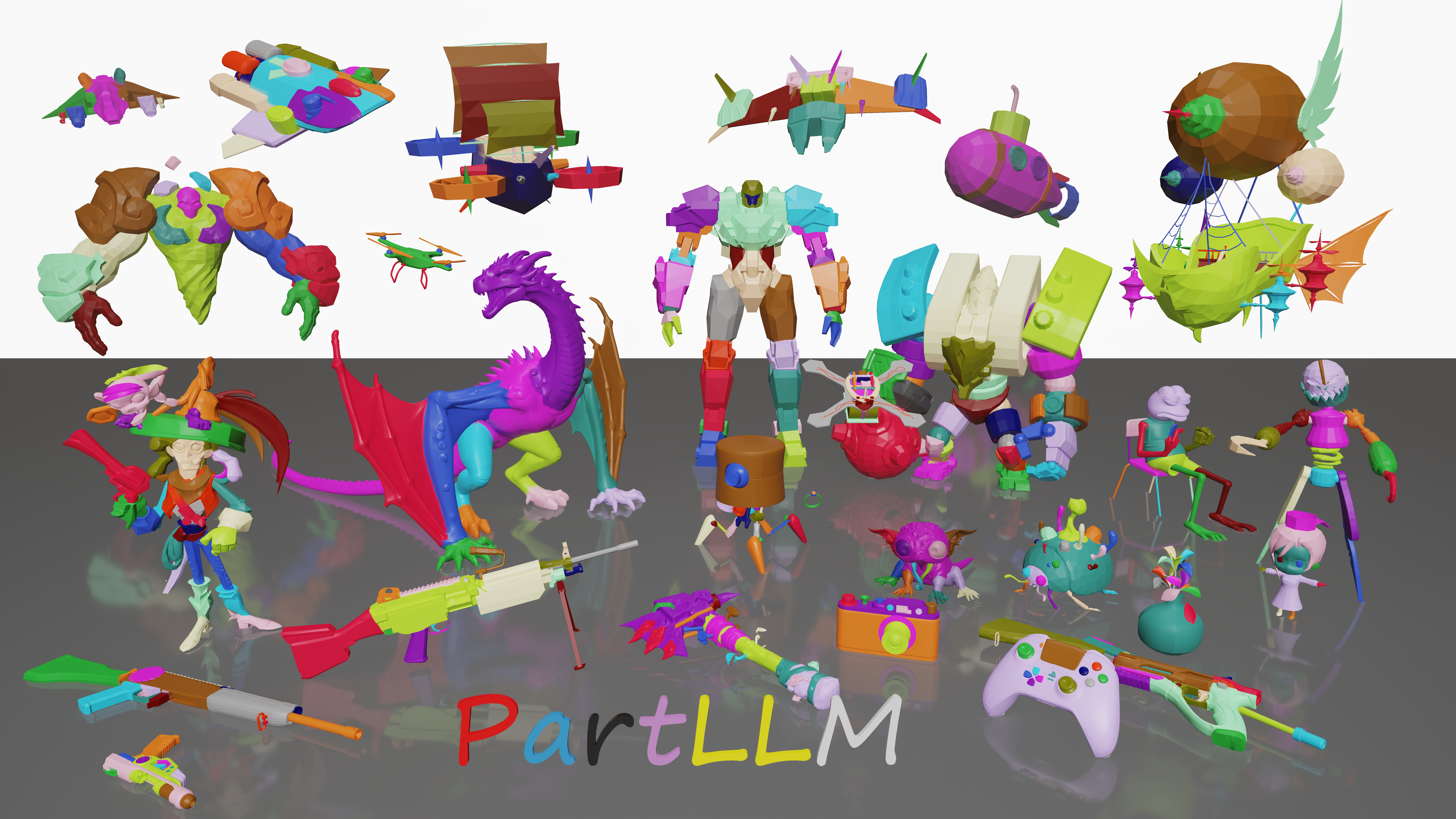}
  \caption{
  Gallery of PartLLM on diverse 3D shapes.
  Across complex artist-created and AI-generated shapes, PartLLM produces semantically meaningful part segmentation with coherent functional structure, clean boundaries, and fine geometric details.
  }
  \Description{Gallery of PartLLM results on diverse 3D objects, showing semantic part decompositions across artist-created and AI-generated assets.}
  \label{fig:ability}
\end{figure*}

\end{abstract}

\ccsdesc[500]{Computing methodologies~Shape analysis}

\keywords{3D part segmentation, open-world segmentation, multimodal large
language models, shape decomposition, geometric understanding}

\begin{teaserfigure}
  \centering
  \includegraphics[width=\textwidth]{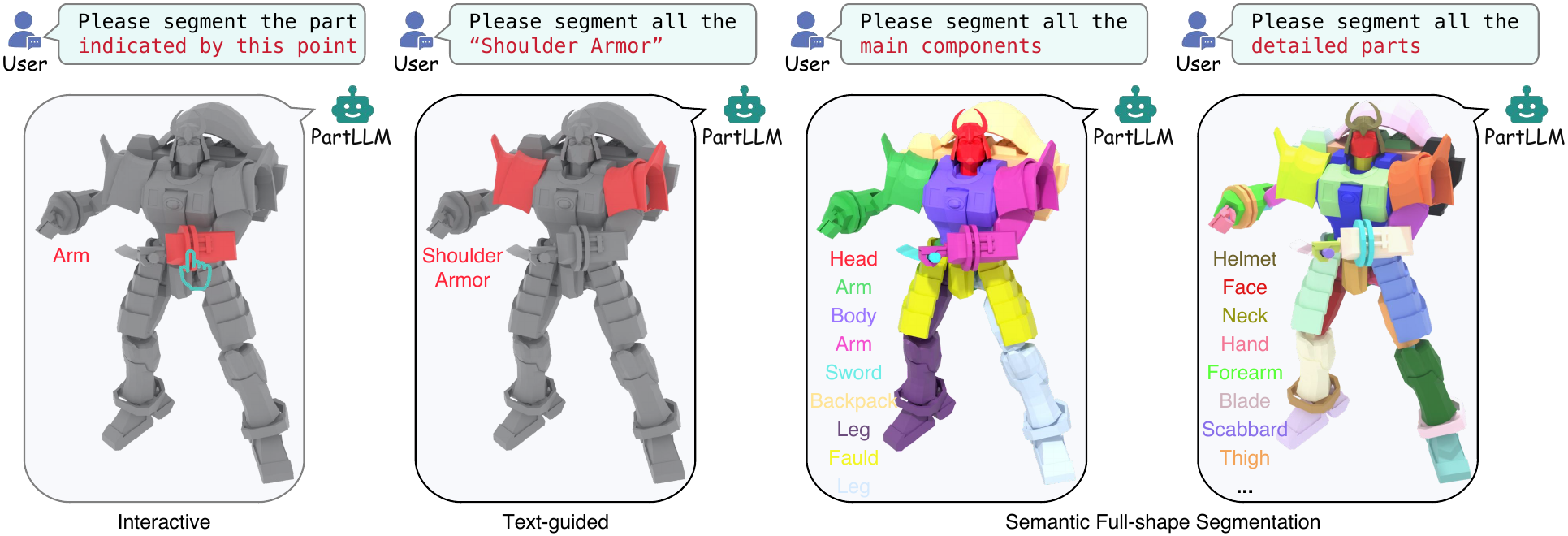}
  \caption{
  PartLLM achieves unified part segmentation within a single model.
  Given the same shape, it produces different segmentations conditioned on
  user intent, including text-guided part segmentation, interactive
  segmentation, and semantic full-shape decomposition at controllable
  granularities.
  }
  \Description{Teaser figure showing one 3D shape segmented by PartLLM under different user intents, including text-guided, point-prompted, and full-shape semantic decomposition settings.}
  \label{fig:teaser}
\end{teaserfigure}

\maketitle

\section{Introduction}

Part-level understanding is fundamental to how humans and machines reason about, create, and interact with 3D objects.
Whether editing a digital asset, rigging a character, or planning a robotic grasp, the first step is almost always identifying what parts the object is made of and how they relate.
Part segmentation makes this understanding computational by decomposing raw 3D geometry into semantically meaningful components.
A long-standing goal is therefore comprehensive 3D part understanding: given a 3D object and a user intent, a model should be able to identify a queried part, support interactive refinement, or decompose the entire shape into semantic parts at a desired granularity.

Despite recent progress, existing methods still treat part segmentation as a collection of separate task settings.
One line of work~\cite{liu2025partfield,zhu2026partsam,p3sam} produces category-agnostic part proposals or full-shape segmentation, but the resulting regions are not tied to explicit semantic identities.
Another line~\cite{ma2024find,jin2026cosmo3d} grounds parts based on input labels, but assumes that the semantic targets are already known and therefore does not decide how the object should be decomposed as a whole.
As Table~\ref{tab:openworld_tasks} summarizes, representative methods cover only subsets of what comprehensive part understanding needs.

\begin{table*}[ht]
\centering
\caption{
Task coverage of representative part segmentation methods.
Existing methods cover only subsets of promptable and full-shape segmentation capabilities, whereas PartLLM supports point-prompted interaction, text-guided part segmentation, category-agnostic segmentation, and semantic full-shape segmentation within a single unified model.
}
\label{tab:openworld_tasks}
\setlength{\tabcolsep}{6pt}
\renewcommand{\arraystretch}{1.08}
\begin{tabular}{l|cc|cc}
\hline
\multirow{2}{*}{Method} & \multicolumn{2}{c|}{Promptable segmentation} & \multicolumn{2}{c}{Full-shape segmentation} \\
\cline{2-5}
& Interactive Point & Part Text & Category-agnostic & Semantic \\
\hline
PartField~\cite{liu2025partfield} & -- & -- & \cmark & -- \\
PartSAM~\cite{zhu2026partsam} & \cmark & -- & \cmark & -- \\
P3-SAM~\cite{p3sam} & \cmark & -- & \cmark & -- \\
S2AM3D~\cite{s2am3d} & \cmark & -- & \cmark & -- \\
SegviGen~\cite{li2026segvigen} & \cmark & -- & \cmark & -- \\
FIND3D~\cite{ma2024find} & -- & \cmark & -- & -- \\
PatchAlign3D~\cite{hadgi2026patchalign3d} & -- & \cmark & -- & -- \\
CoSMo3D~\cite{jin2026cosmo3d} & -- & \cmark & -- & -- \\
\hline
\textbf{Ours} & \cmark & \cmark & \cmark & \cmark \\
\hline
\end{tabular}
\end{table*}

We argue that these tasks should be unified under a common part segmentation formulation.
At their core, these tasks all require inferring a part-level decomposition from a shape under a particular user intent.
The difference lies mainly in how the intent is expressed and what granularity of decomposition is expected.
Moreover, different datasets and interaction modes often provide complementary observations of the same underlying part organization: some reveal only queried parts, some provide unlabeled regions, and others annotate complete semantic decompositions. 
Treating them as separate tasks prevents knowledge learned in one setting from benefiting another. 
A unified formulation is therefore not merely a convenient interface. It provides a scalable route toward comprehensive 3D part understanding by converting heterogeneous task-specific annotations into shared training signals for learning generalizable part representations.

The remaining question is how we should design such a unified 3D segmentation framework.
While most existing approaches adopt a deterministic segmentation model with feature clustering or classification, we advocate that such a unified 3D segmentation should be a conditional generative model.
Unified part segmentation is inherently ambiguous: the number of valid parts varies
across objects, and the same object can be decomposed differently under
different semantic viewpoints or requested granularities.
This ambiguity means part segmentation is not a discriminative task with a fixed answer, but a generative problem conditioned on the intent of the task. Modeling it as a deterministic framework usually results in an averaged and blurry result, while generative frameworks have the potential to sharply segment all parts with some randomness.

Based on this view, we formulate 3D part segmentation as \emph{autoregressive semantic decomposition}, using the
language interface of an LLM to generate a sequence of part representations.
This formulation provides three advantages.
First, it unifies different task settings within the same input-output space: different prompts condition the same generation process, while the generated sequence can contain a variable number of parts at different granularities.
Second, it naturally aligns part segmentation with language modeling, allowing open-vocabulary part names to be generated together with the segmentation rather than being treated as external labels attached after mask prediction.
Third, it makes part prediction contextual: each generated part is conditioned on the parts generated so far, helping the model maintain coherent semantic identities, compatible boundaries, and consistent granularity across the whole shape.

Turning this formulation into an effective segmentation model, however, is non-trivial.
It raises two questions: how to represent each part within a language sequence, and how to convert the generated parts into mutually consistent 3D masks.
We instantiate this formulation as \textbf{PartLLM}, a unified multimodal framework that makes autoregressive semantic decomposition executable.

To address the first question, PartLLM builds a 3D-aware multimodal large language model (MLLM) to express each decomposed part as a hypothesis.
Motivated by the emerging visual perception capabilities of recent MLLMs~\cite{hurst2024gpt,qwen3vl}, we express each part hypothesis as a language-form representation that specifies the part's semantic label, coarse 3D location, and relation to the previously predicted parts.
In this way, the generated part hypotheses define a dynamic, open-vocabulary label set tailored to the current shape and user intent.

To address the second question, we design a decomposition-aware mask decoder to assign 3D points to the generated label set.
A straightforward solution is to use SAM-style decoders from prior work~\cite{zhou2025pointsam,zhu2026partsam} that predict one binary mask per query.
However, by predicting each part as an independent binary mask, such decoders provide limited part-to-part interaction and often lead to overlapping assignments and inconsistent boundaries.
Instead, PartLLM organizes the part hypotheses and the remaining region into a decomposition-aware query set, built from two types of special tokens.
Given this query set, the proposed decoder jointly assigns points to one of the generated part hypotheses or background.
By turning mask prediction into a joint assignment problem, different parts compete within the same decomposition, producing mutually exclusive and semantically coherent masks.
Together with autoregressive generation, this strategy helps maintain consistent identities for symmetric or structurally related parts while producing compatible boundaries across the shape.

PartLLM unifies text-guided segmentation, interactive segmentation, and full-shape segmentation within a single framework.
It can take an arbitrary 3D object as input and directly produce a complete, semantically labeled part decomposition at a specified granularity.
Crucially, the same formulation allows annotations from different tasks and datasets to train a single model, scaling the supervision available for learning 3D parts.
By expressing different tasks and granularities within the same prompt-response format, PartLLM turns 323K unique 3D shapes into 1.1M unified training samples.
Extensive experiments show that it surpasses task-specific methods by large margins across all evaluated settings.
Our contributions are summarized as follows.
\begin{itemize}
\item We reformulate 3D part segmentation as autoregressive semantic decomposition, casting diverse part segmentation tasks as a unified framework.
\item We implement this formulation as PartLLM, where a 3D-aware multimodal language model generates language-form part hypotheses that define a dynamic open-vocabulary label set for the current user intent.
\item We design a decomposition-aware decoding strategy that jointly assigns each input point to predicted part hypotheses, producing globally consistent masks.
\item We demonstrate that the unified formulation enables scalable multi-task training and achieves state-of-the-art performance across all evaluated settings.
\end{itemize}

\section{Related Work}

\subsection{3D Part Segmentation}

\paragraph{Closed-world Part Segmentation}
Early learning-based methods typically cast 3D part segmentation as point-wise classification on 3D shapes~\cite{qi2017pointnet++,thomas2019kpconv,dgcnn,zhao2021point}.
These methods are commonly trained and evaluated on fixed-taxonomy datasets such as ShapeNetPart~\cite{shapenetpart} and PartNet~\cite{mo2019partnet}.
This closed-world assumption limits their applicability to real 3D assets, which often contain object categories and part labels outside the predefined training taxonomy.

\paragraph{Lifting 2D Foundation Models}
To reduce the dependence on closed 3D taxonomies, recent methods transfer priors from 2D foundation models, including vision--language models and image segmentation models~\cite{clip,li2022grounded,oquab2024dinov2,sam1}.
One line of work performs text-driven 3D part segmentation by matching rendered views or 3D regions with part names in a vision--language feature space~\cite{abdelreheem2023satr,liu2023partslip,zhu2023pointclip,garosi20253d}.
Another line lifts mask predictions from image segmentation models into 3D, either by merging multi-view masks or by distilling 2D mask features into 3D representations~\cite{zhou2023partslip++,yang2024sampart3d,zhong2024meshsegmenter,tang2024segment,lang2024iseg,xue2025zerops}.
While these methods bring strong 2D priors to 3D part segmentation, they often depend on rendering, view aggregation, or per-shape processing, making the resulting segmentation sensitive to view coverage and costly to apply at scale.

\paragraph{Feed-forward 3D Models}
Recent feed-forward models avoid per-shape lifting by predicting part masks directly from 3D inputs.
One line of work follows a semantic grounding formulation.
FIND3D~\cite{ma2024find} pioneered this direction by training a point cloud network with text-aligned features, allowing users to query and localize target parts on 3D shapes using free-form text descriptions.
PatchAlign3D~\cite{hadgi2026patchalign3d} and CoSMo3D~\cite{jin2026cosmo3d} follow this route, improving semantic grounding through local region alignment and canonical spatial modeling, respectively.
Another line of work focuses on category-agnostic part decomposition.
PartField~\cite{liu2025partfield} initiated this direction by learning a part-aware 3D feature field and clustering it into hierarchical full-shape decompositions.
Motivated by SAM, subsequent methods such as PartSAM~\cite{zhu2026partsam}, P3-SAM~\cite{p3sam}, and S2AM3D~\cite{s2am3d} develop promptable 3D models for category-agnostic mask prediction.
More recently, SegviGen~\cite{li2026segvigen} takes a different route by repurposing a 3D generative model and treating segmentation as coloring a shape.
Name That Part~\cite{paul2025name} introduces ALIGN-Parts, which combines geometric and multi-view appearance features to predict part-level representations and matches them to affordance-aware text descriptions, producing a complete named part decomposition.
Taken together, these feed-forward models make open-world 3D part segmentation more scalable and controllable, but they each address only a subset of part segmentation tasks.
Text-based methods attach semantics to geometry, yet they rely on user-provided part labels and cannot produce a full decomposition autonomously.
Category-agnostic methods can decompose the full shape, yet their outputs remain geometric masks without semantic identities.
Our work unifies all these capabilities under a single formulation by treating part segmentation as conditional sequential generation, enabling the first model that handles semantic grounding, point-based interaction, and semantic full-shape decomposition together.

\subsection{MLLMs for Visual Perception}

\paragraph{MLLMs for 2D Perception}
Recent MLLMs have extended visual instruction following from image-level responses to spatially grounded perception.
One line of work represents image regions through coordinate tokens or region features, enabling language models to refer to, describe, and localize visual entities with boxes or regions~\cite{peng2023kosmos,chen2023shikra,you2023ferret,bai2023qwenvl,rexomni}.
Another line further connects language models with dense segmentation, enabling pixel-level referring and reasoning segmentation in 2D images~\cite{lai2024lisa,rasheed2024glamm,ren2024pixellm}.
These methods often introduce special tokens whose hidden representations act as mask-level handles, bridging language generation and dense visual prediction.
In 3D shapes, however, parts are not isolated regions but compositional units: their identities and boundaries are determined jointly by the whole shape and other parts.
Although our method shares a similar idea of using special tokens to condense information from LLM outputs, each token in PartLLM represents a part hypothesis within an object-level decomposition rather than a standalone target handle.
We further design a decoding strategy where all generated hypotheses are grounded together, so that masks are mutually consistent and reflect the structure of the generated decomposition.

\paragraph{MLLMs for 3D Perception}
Recent multimodal language models extend language-based reasoning to the 3D domain by aligning 3D representations with large language models.
Early systems support 3D captioning, question answering, and grounding by injecting scene features or object-centric representations into language models~\cite{hong2023threedllm,chen2023ll3da,huang2023chatscene}.
More recent models make spatial prediction more explicit, using language models to reason about or directly generate 3D boxes, object identifiers, and structured scene layouts~\cite{cho2024cubellm,zhu2024llava3d,mao2025spatiallm}.
A recent part-aware 3D MLLM further explores structured program generation for part-based reasoning, generation, and editing~\cite{wang2025partxmllm}.
However, such program-level outputs remain symbolic and coarse; they do not provide dense part segmentation masks that organize the surface geometry into a coherent part decomposition.
In contrast, PartLLM bridges the generative capacity of language models with dense geometric prediction, using autoregressive generation not as a reasoning interface but as the decomposition mechanism itself, directly producing point-level masks tied to the generated semantic structure.

\section{Method}
\label{sec:method}

\begin{figure*}[t]
  \centering
  \includegraphics[width=\textwidth]{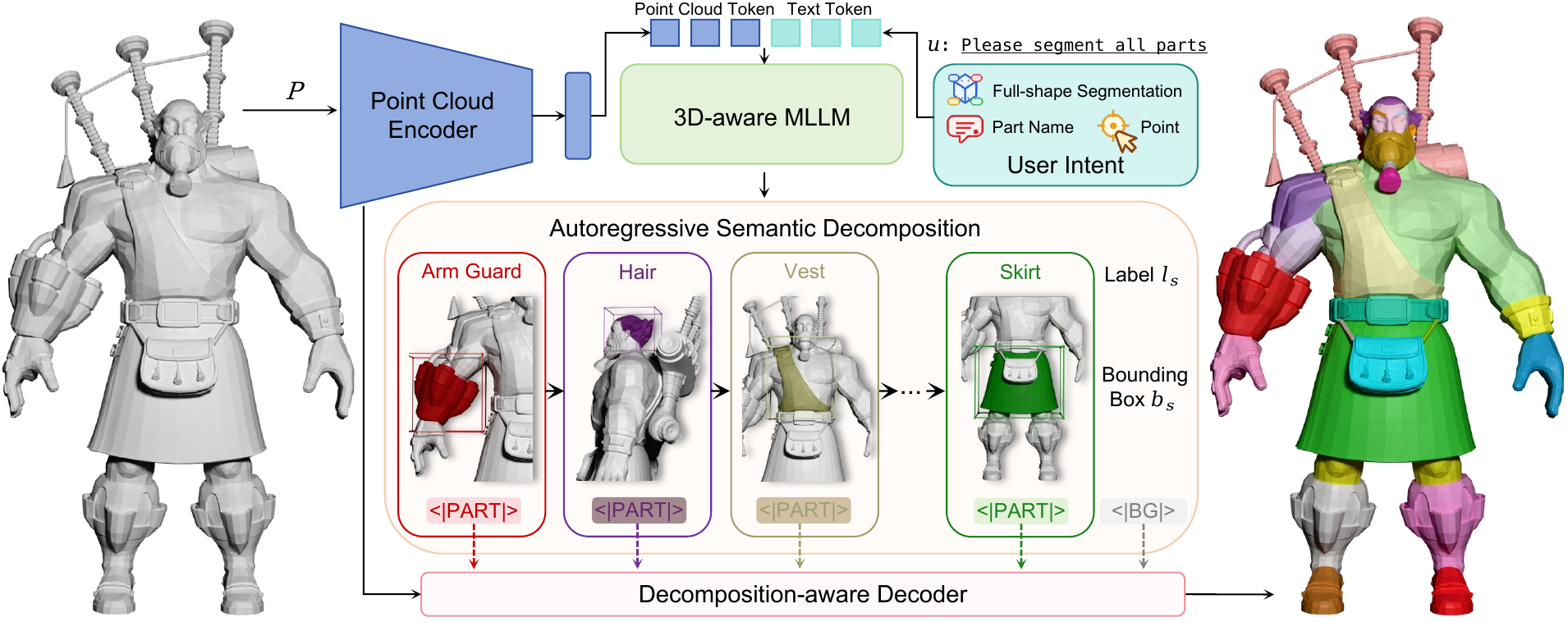}
  \caption{
  Overview of PartLLM.
  Different part segmentation tasks are unified as autoregressive semantic decomposition: the user intent (right) conditions a 3D-aware multimodal language model to generate task-specific part hypotheses, which are then jointly decoded into mutually consistent masks.
  }
  \Description{Overview of the PartLLM pipeline. A point cloud encoder produces
  3D tokens for a multimodal language model conditioned on user intent. The
  model generates named and localized part hypotheses, which a
  decomposition-aware decoder converts into a complete colored segmentation.}
  \label{fig:overview}
\end{figure*}

Fig.~\ref{fig:overview} gives an overview of PartLLM.
Given an input point cloud $P=\{p_i\}_{i=1}^{N}$ sampled from a 3D shape and a user intent $u$, such as ``\texttt{Please segment all parts}'' for the full-shape segmentation task, PartLLM first uses a 3D-aware MLLM to autoregressively generate a sequence of part hypotheses $Y=(y_1,\ldots,y_S)$.
Each hypothesis $y_s=(l_s,b_s)$ specifies a part label $l_s$ together with a coarse 3D location $b_s$, indicating what part should be segmented and where it is roughly located.
Conditioned on these generated hypotheses, a decomposition-aware decoder then assigns each point to one of the hypotheses or the remaining region, producing a dense assignment $q$ and the corresponding point-level masks.

\subsection{3D-Aware MLLM for Part Hypothesis Generation}
\label{sec:method-generator}

\paragraph{3D Tokenization.}
To generate the part hypothesis sequence based on the input geometry, we attach a point cloud encoder~\cite{ptv3} to the input token stream of an MLLM~\cite{qwen3vl}.
Given the input point cloud $P$, the encoder first extracts full-resolution
point features for all input points.
We then downsample the point cloud to $K$ center points and gather their
corresponding encoder features.
A linear projector maps these sampled point features into the language hidden
space, where they are inserted at the \texttt{<point\_cloud>} positions and
serve as 3D visual tokens for the MLLM.
The full-resolution point features are retained
for the mask decoder, so the language model receives a compact token sequence
while the decoder still has access to dense geometric evidence. 

\paragraph{Hypothesis Definition.}
The MLLM generates the hypothesis sequence $Y=(y_1,\ldots,y_S)$ in an autoregressive text format.
Each hypothesis $y_s=(l_s,b_s)$ is serialized as a structured text span containing a semantic label $l_s$ and an axis-aligned 3D box $b_s$, followed by a special token \texttt{<|PART|>}.
For example, a generated hypothesis may take the form
\texttt{label=Backrest, bbox=[...] <|PART|>}.
Here, \texttt{bbox} denotes an axis-aligned 3D box
$[x_{\min},y_{\min},z_{\min},x_{\max},y_{\max},z_{\max}]$ that serves as a
spatial cue.
The \texttt{<|PART|>} token acts as the terminal token of a part
decision, rather than a generic separator.
After the language model forward pass, the hidden state at each generated
\texttt{<|PART|>} token is used as the representation of the corresponding part
hypothesis.
Through attention, the \texttt{<|PART|>} token summarizes the generated label,
part location, user intent, input $P$, and previously predicted parts
into a context-dependent representation of the part decision.
This representation encodes what the part is, where it is, and how it relates
to the current decomposition state.
In addition to the \texttt{<|PART|>} tokens, the assistant response ends with a special
\texttt{<|BG|>} token.
Unlike the \texttt{<|PART|>} tokens, \texttt{<|BG|>} does not correspond to any
generated part.
Instead, its hidden state represents the context-dependent complement of the
generated parts.
Together, the \texttt{<|PART|>} token states and the \texttt{<|BG|>} token state serve as decoder queries for mask generation, as described next.

\subsection{Decomposition-Aware Decoder}
\label{sec:method-decoder}

The generated token states are used as dynamic queries for mask decoding.
Let $h_s$ denote the hidden state of the $s$-th \texttt{<|PART|>} token, and let
$h_{S+1}=h_{\mathrm{bg}}$ denote the hidden state of \texttt{<|BG|>}.
These states form the decomposition query set:
\begin{equation}
  H=[h_1,\ldots,h_S,h_{\mathrm{bg}}].
\end{equation}
Let $X$ denote the full-resolution point features provided by the point cloud encoder.
Given $H$ and $X$, the decoder first uses a two-way transformer~\cite{sam1} to contextualize the queries and point features jointly:
\begin{equation}
  (\tilde{H}, \Phi) = T_\psi(H, X),
\end{equation}
where $\tilde{H}=[\tilde{h}_1,\ldots,\tilde{h}_{S+1}]$ are the updated query
features and $\Phi=\{\phi_i\}_{i=1}^{N}$ denotes the point-wise geometric
features used for mask prediction.
This query-point interaction allows each generated part hypothesis to attend to the input geometry while injecting the current decomposition context into point-wise features.

For each contextualized query $\tilde{h}_s$, $s\in\{1,\ldots,S+1\}$, an MLP-based hypernetwork $\rho_\theta$ predicts a query-specific classifier $w_s$.
The compatibility between query $s$ and point $p_i$ is then computed as
\begin{equation}
  w_s=\rho_\theta(\tilde{h}_s), \qquad
  z_{i,s}=w_s^\top \phi_i .
\end{equation}
For each point, this produces a logit vector
$z_i\in\mathbb{R}^{S+1}$, where each dimension corresponds to one generated
part hypothesis or the background query.
The decoder predicts a categorical distribution over the generated query set:
\begin{equation}
  p_\theta(q_i=s \mid P,u,Y)
  =
  \frac{\exp(z_{i,s})}{\sum_{r=1}^{S+1}\exp(z_{i,r})},
  \quad s\in\{1,\ldots,S+1\}.
\end{equation}
Here, $q_i\in\{1,\ldots,S+1\}$ is the assignment variable of point $p_i$.
For $s\leq S$, $q_i=s$ indicates that $p_i$ belongs to the $s$-th generated part hypothesis, while $q_i=S+1$ assigns the point to the background.
The mask of each generated part is therefore induced by the points assigned to its corresponding hypothesis.

\paragraph{Discussion.}
The key distinction from SAM-style decoders~\cite{zhu2026partsam,p3sam} lies in the prediction space.
Given $S$ generated hypotheses, a SAM-style decoder would treat them as $S$ independent binary mask predictions.
PartLLM instead treats the generated hypotheses and the background query as a dynamic label set, and predicts an $(S+1)$-way assignment for each point.
This is possible because the ambiguity of what to segment has already been resolved before mask decoding: the MLLM generates an explicit hypothesis sequence $Y$, leaving the decoder to ground this generated label set into dense masks.
The shared categorical distribution explicitly couples all masks by forcing the generated parts and background to compete for the same points, yielding mutually exclusive assignments.
This joint assignment is especially important for part segmentation, where neighboring or structurally related parts often have ambiguous boundaries and should be interpreted relative to each other within the same shape-level decomposition.

\begin{figure}[ht]
  \centering
  \includegraphics[width=\columnwidth]{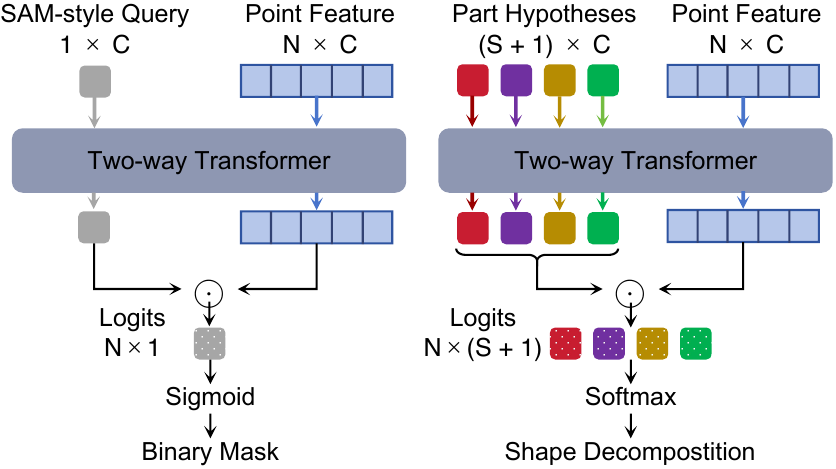}
  \caption{
  Comparison of two mask decoding paradigms.
  SAM-style decoders predict each mask independently, using a sigmoid decision
  for every query.
  PartLLM instead performs decomposition-aware joint decoding.
  Autoregressively generated part hypotheses form a dynamic query set, and a softmax competition assigns each point
  to one generated part or background.
  This converts mask prediction from independent binary decisions into a joint
  decomposition of the shape.
  }
  \Description{Diagram of the PartLLM decoder. Generated part and background
  queries interact with point features, produce an N by S plus one score
  matrix, and use row-wise softmax for joint point assignment. A small
  SAM-style reference indicates independent binary mask decoding with sigmoid.}
  \label{fig:decoder}
\end{figure}

\subsection{Training Objective}
\label{sec:method-training}

Training supervises both parts of the representation: the autoregressive
hypothesis sequence and the dense point assignment.
Let $a^\star$ denote the target assistant response that serializes the part
labels, coarse 3D locations, and special tokens.
The language model is trained with the standard next-token objective:
\begin{equation}
  \mathcal{L}_{\mathrm{lm}}
  =
  -\sum_t \log p_\theta(a^\star_t \mid P,u,a^\star_{<t}).
\end{equation}
This loss teaches the model to generate the semantic decomposition in language
form, including the \texttt{<|PART|>} and \texttt{<|BG|>} tokens whose hidden
states are used by the decoder.

For mask supervision, let $Z=\{z_{i,s}\}\in\mathbb{R}^{N\times(S+1)}$ denote
the decoder logits over the $S$ generated part hypotheses and the background
query.
The target assignment $q^\star\in\{1,\ldots,S+1\}^{N}$ follows the response
order: points belonging to the $s$-th target part are assigned to
class $s$, and points outside the target parts are assigned to class
$S+1$. 
The mask loss is a point-level cross-entropy over this dynamic label space:
\begin{equation}
  \mathcal{L}_{\mathrm{mask}}
  =
  -\frac{1}{N}\sum_{i=1}^{N}
  \log
  \frac{\exp(z_{i,q^\star_i})}
  {\sum_{s=1}^{S+1}\exp(z_{i,s})}.
\end{equation}
Together with the decomposition-aware decoding strategy,
this target construction turns mask supervision into a mutually exclusive point
classification problem over the generated hypotheses.
Unlike independent binary mask losses, each point contributes to exactly one
class in the dynamic label space, forcing the generated parts and the background
query to compete during training.
As a result, the generated part hypotheses are optimized as a coupled
decomposition of the shape, where the assignment of one part is learned relative
to the others.

The total training objective is a weighted combination of the sequence and mask losses:
\begin{equation}
  \mathcal{L}
  =
  \mathcal{L}_{\mathrm{lm}}
  +
  \lambda_{\mathrm{mask}}\mathcal{L}_{\mathrm{mask}}.
\end{equation}

\subsection{Unified Task Formulation}
\label{sec:method-tasks}

\begin{table*}[ht]
\centering
\footnotesize
\caption{
Prompt-response template for the unified task formulation.
}
\label{tab:task_io}
\setlength{\tabcolsep}{4pt}
\renewcommand{\arraystretch}{0.91}
\begin{tabular}{@{}p{0.11\textwidth}p{0.53\textwidth}p{0.32\textwidth}@{}}
\toprule
Task & Prompt input & Generated response \\
\midrule
Full-shape &
\parbox[c]{0.50\textwidth}{\ttfamily\raggedright
general: Please segment and name all parts in <point\_cloud>.\par
coarse: Please segment and name the main parts in <point\_cloud>.\par
fine: Please segment and name all detailed parts in <point\_cloud>.\par
number: Please segment and name about \{n\} parts in <point\_cloud>.} &
\begin{tabular}[c]{@{}l@{}}
\texttt{label=Backrest, bbox=[...] <|PART|>}\\
\texttt{label=Seat, bbox=[...] <|PART|>}\\
\texttt{label=Armrest, bbox=[...] <|PART|>}\\
\texttt{label=Base, bbox=[...] <|PART|>}\\
\texttt{<|BG|>}
\end{tabular} \\
\midrule
Text-guided &
\texttt{Please segment the Backrest and Armrest in <point\_cloud>.} &
\begin{tabular}[c]{@{}l@{}}
\texttt{label=Backrest, bbox=[...] <|PART|>}\\
\texttt{label=Armrest, bbox=[...] <|PART|>}\\
\texttt{<|BG|>}
\end{tabular} \\
\midrule
\multirow{2}{=}{Interactive} &
\texttt{First click: segment the part at (x, y, z) in <point\_cloud>.} &
\texttt{label=Armrest, bbox=[...] <|PART|> <|BG|>} \\
\cmidrule(l){2-3}
&
\texttt{Refinement: previous mask + include/exclude point (x, y, z).} &
\texttt{label=Armrest, bbox=[...] <|PART|> <|BG|>} \\
\bottomrule
\end{tabular}
\end{table*}

All part segmentation tasks reduce to the same problem: determining which parts to produce and where, conditioned on the user intent.
The autoregressive semantic decomposition formulation reflects this shared
structure by fixing the response format as a sequence of generated part
hypotheses followed by \texttt{<|BG|>}.
Here, \texttt{<|BG|>} represents the context-dependent complement of the
generated hypotheses, rather than a fixed background category.
Task differences are encoded in the input prompt, which determines both the
generated part set and the meaning of its complement.
By always including \texttt{<|BG|>} in the query set, the decoder performs the same softmax assignment over all points regardless of how many parts are generated, eliminating the need for task-specific decoding logic.
This context-dependent behavior emerges from training with the same cross-entropy loss across all modes, allowing \texttt{<|BG|>} to learn a consistent role as the complement of whatever the current prompt requests.
Table~\ref{tab:task_io} summarizes the prompt-response interface for each task mode.
\begin{itemize}
\item \textbf{Full-shape segmentation.}
The general full-shape prompt asks the model to generate all parts of the
object.
We further control the granularity with coarse, fine, and number-based
prompt variants, so the same model can produce decompositions at different
levels of detail.
Besides this general open-ended prompt, these controlled prompt types provide
an explicit interface for requesting different decomposition granularities.
The construction details of these prompt variants are described in
Sec.~\ref{sec:exp-setup}.
In this setting, \texttt{<|BG|>} handles uncovered or unlabeled regions.

\item \textbf{Text-guided part segmentation.}
The prompt provides a list of queried part names, and the model generates only
the requested parts, while \texttt{<|BG|>} absorbs non-queried parts.

\item \textbf{Interactive segmentation.}
The first interaction round provides a single 3D point, and the model generates the
part containing that point.
Subsequent rounds refine the previously predicted mask: we encode the previous
mask as point colors in the input point cloud and provide an include or exclude
point to refine the mask.
In all interaction rounds, \texttt{<|BG|>} covers points outside the target
part.
\end{itemize}

\subsection{Scalable Training with Heterogeneous Data}
\label{sec:method-scalable-training}

Because task differences are expressed entirely through prompts and target
responses, heterogeneous annotations can be converted into the same
supervision format and mixed within one model.
For semantic part annotations, each target hypothesis contains the annotated
part label, its coarse 3D location, and a \texttt{<|PART|>} token.
For category-agnostic decompositions, we retain the same response structure
and simply use a generic label such as \texttt{part}.
From the same annotated shape, we derive full-shape segmentation training examples by varying the
granularity prompt, text-guided examples by sampling target part subsets, and
interactive examples by sampling target points, with optional refinement
rounds built from perturbed masks.
Regardless of the original annotation type or task mode, we train
the model on all resulting instances using the same autoregressive generation
objective and decomposition-aware decoder. Training data construction details are
given in Sec.~\ref{sec:exp-setup}.

Viewed more broadly, this unified formulation turns task and data diversity into
a scaling axis.
As more datasets become available, the model can incorporate them through prompt and response design alone, without architectural changes or new training objectives.

\begin{figure*}[!t]
  \centering
  \includegraphics[width=0.97\textwidth]{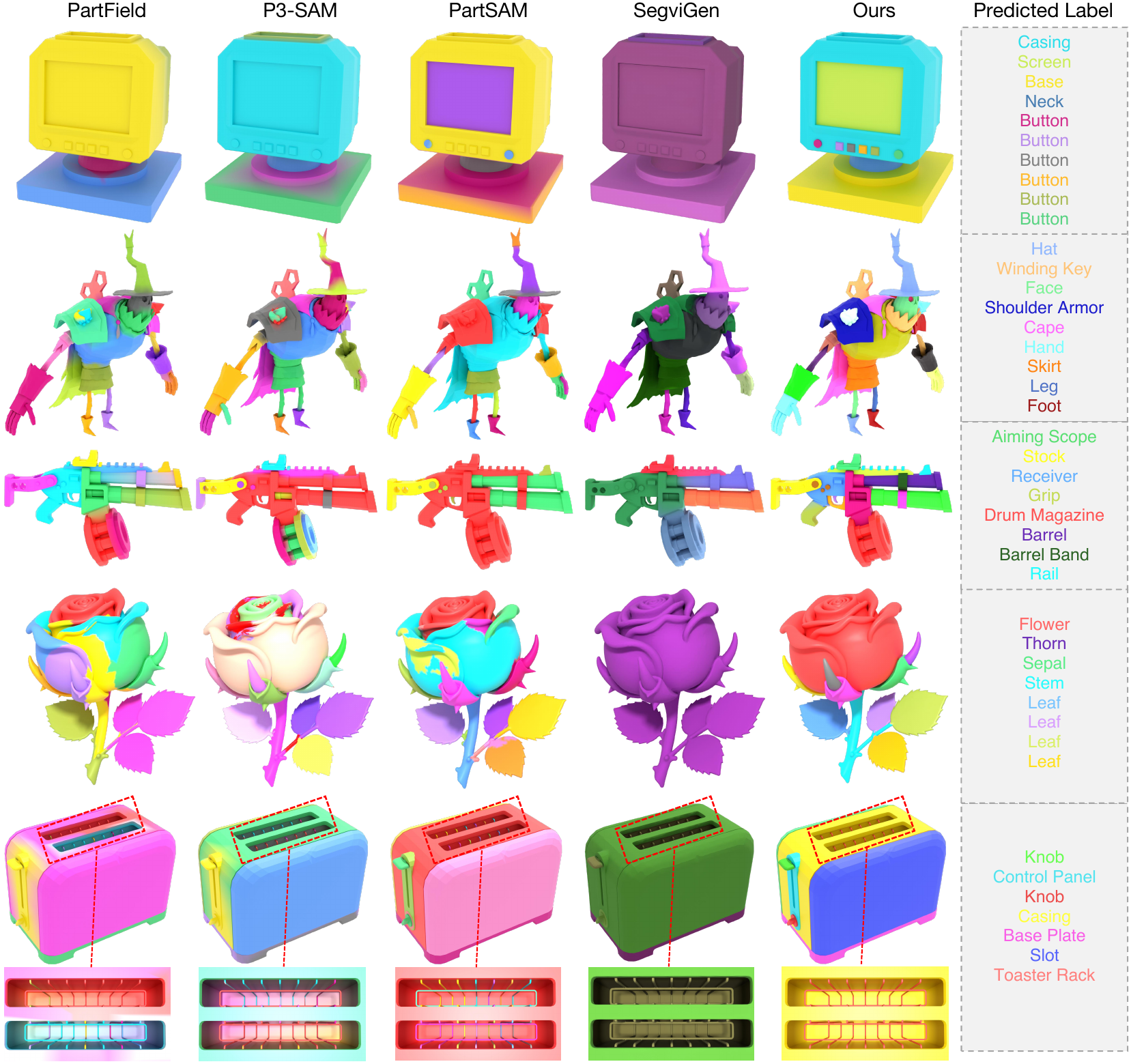}
  \caption{
  Qualitative comparison with baseline methods~\cite{liu2025partfield,p3sam,zhu2026partsam,li2026segvigen} on full-shape segmentation.
  }
  \Description{Comparison of full-shape part decompositions produced by
  PartField, P3-SAM, PartSAM, SegViGen, and PartLLM on five 3D objects.
  Different colors denote the predicted parts.}
  \label{fig:full_comp}
\end{figure*}

\section{Experiments}
\label{sec:experiments}

We evaluate PartLLM on three 3D part segmentation task families within our
open-world formulation: full-shape segmentation, text-guided part
segmentation, and interactive segmentation.

\begin{figure}[ht]
  \centering
  \includegraphics[width=\columnwidth]{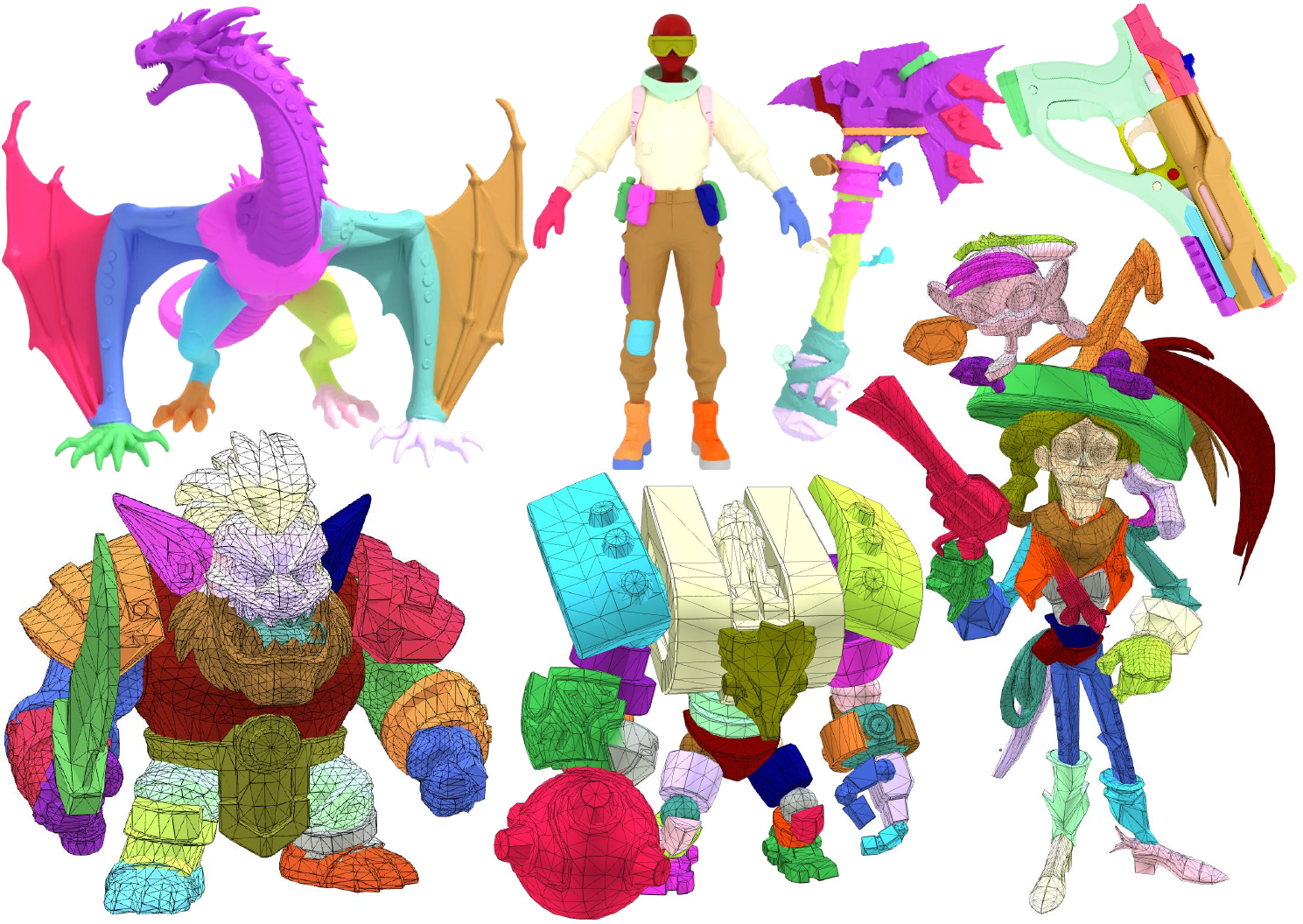}
  \caption{
  Qualitative full-shape segmentation results on AI-generated meshes.
  The first row contains dense meshes produced by geometry-based generative models~\cite{hunyuan3d,trellis2}, while the second row contains low-poly meshes produced by autoregressive topology generative models~\cite{zhao2025deepmesh,liu2026quadgpt}.
  }
  \Description{Colored full-shape part segmentations of dense and low-poly
  AI-generated 3D meshes, including characters, creatures, and objects.}
  \label{fig:aigc_high}
\end{figure}

\subsection{Setup}
\label{sec:exp-setup}

\paragraph{Training Data.}
We train PartLLM on a large-scale mixture of 3D part segmentation
datasets.
We first curate three public datasets with semantic part labels:
PartNeXt~\cite{wang2026partnext}, 3DCoMPaT200~\cite{3dcompat200}, and
PartVerse~\cite{ding2026fullpart}.
We then use an MLLM to annotate 75K licensed 3D assets with semantic part
labels.
Together, these data provide full-shape, text-guided, and interactive
supervision under a shared prompt-response interface.
We additionally use a training subset of
HY3D-Bench~\cite{hunyuan3d2026hy3dbench} for category-agnostic geometric
decomposition without semantic part names.
For each shape, we construct training instances across all applicable task
modes and granularity levels, yielding 1.1M training samples in total from
323K unique shapes.
For full-shape supervision, granularity is defined according to the annotation
structure.
For the hierarchical datasets PartNeXt and 3DCoMPaT200, the shallowest valid
decomposition is treated as coarse, the deepest valid decomposition is treated
as fine unless it contains very few parts, and intermediate levels are grouped
by part count, with at most 6 parts as coarse, 7--12 as medium, and 13 or
more as fine.
For all other datasets, granularity is determined directly by the same
part-count thresholds.
We then sample the full-shape prompt from a small family of general,
coarse/fine, and number-controlled templates.

\paragraph{Benchmarks.}
We evaluate on multiple benchmarks that cover different task settings.
From PartNeXt~\cite{wang2026partnext}, we evaluate on 500 shapes covering 50 object
categories.
From 3DCoMPaT200~\cite{3dcompat200}, we use about 2,000 shapes across 200
categories.
Both datasets provide multi-level part annotations, allowing us to evaluate
whether a model can handle different semantic granularities across categories.
From HY3D-Bench~\cite{hunyuan3d2026hy3dbench}, we select 100 shapes with a
roughly uniform distribution over part counts.
This benchmark is challenging, with the most complex case containing up to 50
ground-truth parts.
For full-shape segmentation, we additionally evaluate on PartObjaverse-Tiny~\cite{yang2024sampart3d}, a common
benchmark for category-agnostic part segmentation.
For text-guided part segmentation, we additionally use PartNet-E~\cite{liu2023partslip}.
All evaluation shapes are strictly held out from the training set, and all
quantitative results are reported using mIoU-based metrics unless otherwise
specified.

\paragraph{Implementation details.}
Our 3D-aware MLLM is built on Qwen3-VL-4B~\cite{qwen3vl}.
The point cloud encoder is a PointTransformerV3~\cite{ptv3} initialized from
Utonia~\cite{zhang2026utonia} pretrained weights, which takes an input point
cloud of 81,920 points and produces per-point features of dimension 1386.
We then select 4,096 points via farthest-point sampling and project their
features to the language model's hidden dimension through a linear layer.
In the mask decoder, we use a two-way transformer with two layers, an
embedding dimension of 512, and eight attention heads.
Each layer contains self-attention on query tokens, cross-attention from
queries to point features, an MLP, and cross-attention from point features
back to queries.
This transformer operates on the 4,096 downsampled point features together
with the part-hypothesis hidden states.
The updated point features are then interpolated back to the original point
resolution and merged with the encoder's full-resolution features to obtain
the point-wise decoder features $\phi_i$ used in Sec.~\ref{sec:method}.
For autoregressive training, we serialize the target part hypotheses in
ascending order of their bounding-box center coordinates along the $x$, $y$,
and $z$ axes.
We apply random 3D rotation augmentation to each training sample. Bounding-box
and point prompts are generated after augmentation in the transformed
coordinate frame.
We train all modules end-to-end on 48 H20 GPUs for 5 epochs with AdamW
($\beta_1{=}0.9$, $\beta_2{=}0.95$, weight decay 0.02), using 2\% warmup
followed by cosine learning rate decay to 0.1$\times$ the initial learning
rate.
All components are jointly optimized with differentiated learning rates:
$1{\times}10^{-5}$ for the point cloud encoder,
$5{\times}10^{-5}$ for projection layers, $1{\times}10^{-5}$ for the LLM,
and $1{\times}10^{-4}$ for the mask decoder.
The mask loss weight $\lambda_{\mathrm{mask}}{=}1.0$.

\subsection{Evaluation on Full-Shape Segmentation}
\label{sec:exp-full}

We first evaluate the central capability of PartLLM, i.e., decomposing a complete
3D shape into parts.
We study this capability in two settings.
Category-agnostic full-shape segmentation evaluates whether a method can
predict the part structure of the whole shape without predicting the semantic label.
Semantic full-shape segmentation further evaluates whether the method can
produce a complete decomposition with meaningful part labels.

\subsubsection{Category-Agnostic Full-Shape Segmentation}
This protocol measures how well a method can recover the ground-truth part
structure under its best operating granularity.
We compare with PartField~\cite{liu2025partfield},
P3-SAM~\cite{p3sam}, PartSAM~\cite{zhu2026partsam},
SAMPart3D~\cite{yang2024sampart3d},
and SegviGen~\cite{li2026segvigen}.
Following the evaluation protocol of PartField~\cite{liu2025partfield} and
PartSAM~\cite{zhu2026partsam}, we evaluate each method using decompositions
generated at multiple candidate granularities.
For PartLLM, we use the general, coarse, and fine full-shape prompts defined in
Table~\ref{tab:task_io} under a fixed random seed.
We do not use number-controlled prompts in this evaluation and therefore do
not provide the ground-truth part count at test time.
For PartField, we use clustering results with the number of clusters ranging
from 1 to 20.
For PartSAM and P3-SAM, we adjust their post-processing parameters to obtain
three sets of masks at different granularities.
SegviGen does not support explicit multi-granularity control, so we sample
three outputs using different random seeds.
Each ground-truth part mask is then matched to the predicted mask with the
highest IoU across all granularities, and we report the average of these
best-matched IoU values.

\paragraph{Results.}
Table~\ref{tab:agnostic_results} reports the quantitative results.
PartLLM consistently outperforms all baselines across the evaluated benchmarks.
The gains are especially large on 3DCoMPaT200, where PartLLM
improves over the strongest baseline by 34.5 mIoU at the coarse level
and 32.0 mIoU at the fine level.
Fig.~\ref{fig:full_comp} shows qualitative comparisons.
Existing methods often produce decompositions with unstable granularity:
they may merge distinct functional parts, split a single coherent part into
fragments, or miss small structures.
In particular, promptable methods such as PartSAM and P3-SAM can recover many
local regions, but their over-segmented masks are not associated with semantic
part identities and are therefore not always meaningful as object parts.
PartLLM produces more complete and coherent full-shape segmentations, with
part boundaries that better align with the underlying object structure.
Fig.~\ref{fig:aigc_high} further shows examples on AI-generated meshes, where
PartLLM produces plausible part decompositions for outputs from different
generation paradigms, including dense meshes produced by geometry-based
generative models and low-poly meshes produced by autoregressive topology
generative models.

\begin{figure}[ht]
  \centering
  \includegraphics[width=\columnwidth]{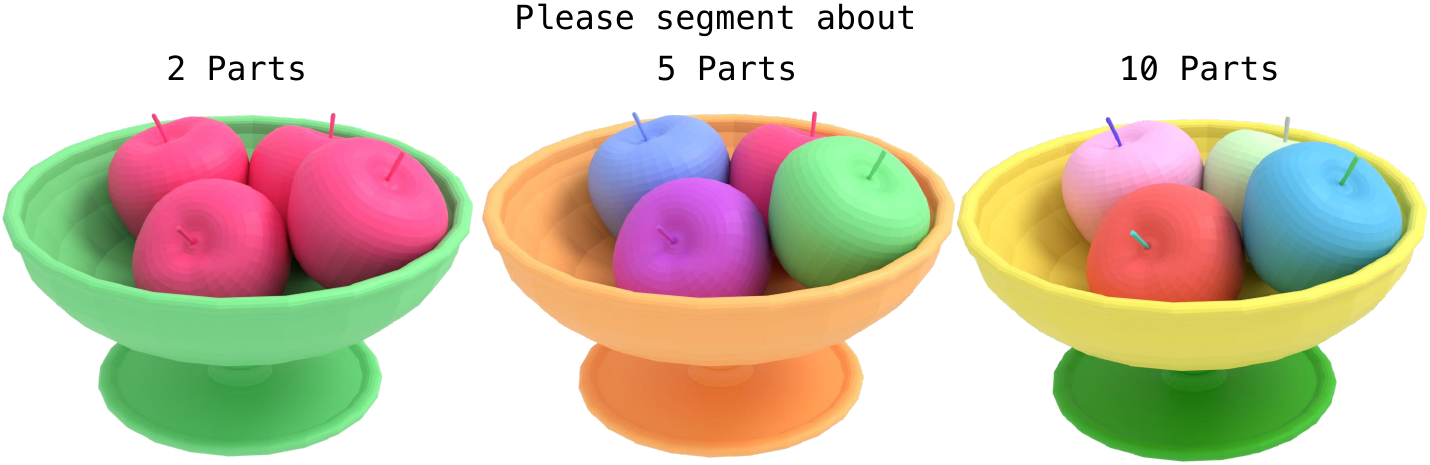}
  \caption{
  Number-based granularity control.
  By changing the requested part counts, PartLLM produces progressively finer
  decompositions while preserving semantic coherence.
  }
  \Description{A number-control figure showing the same object decomposed by PartLLM under prompts requesting about 2 parts, about 5 parts, and about 10 parts.}
  \label{fig:number_control}
\end{figure}

We attribute these advantages to two properties of our formulation.
First, autoregressive generation predicts parts as a coupled decomposition
rather than as independent regions, which leads to more coherent full-shape
segmentations.
Second, granularity control is built into the generation process itself, so the
model can produce decompositions that remain consistent with the requested
level of detail.
Fig.~\ref{fig:number_control} further shows that the same interface also
supports prompt-based number control: as the requested part count increases,
the model produces progressively finer yet still semantically coherent
decompositions.

\begin{table}[ht]
\centering
\caption{Quantitative results on category-agnostic full-shape segmentation.
We report mask-only mIoU by matching each ground-truth part to its best-IoU
prediction across granularities.}
\label{tab:agnostic_results}
\small
\setlength{\tabcolsep}{0pt}
\begin{tabular}{@{}l@{\hspace{6pt}}c@{\hspace{7pt}}c@{\hspace{3pt}}c@{\hspace{7pt}}c@{\hspace{6pt}}c@{}}
\toprule
\multirow{2}{*}{Method} & \multirow{2}{*}{PartNeXt} & \multicolumn{2}{c}{3DCoMPaT200} & \multirow{2}{*}{HY3D-Bench} & \multirow{2}{*}[-0.55ex]{\shortstack{PartObjaverse\\-Tiny}} \\
\cmidrule(lr){3-4}
 &  & Coarse & Fine &  &  \\
\midrule
PartField & 42.7 & 45.8 & 35.0 & 33.6 & 51.5 \\
SAMPart3D & 40.8 &
51.6 & 41.1 & 36.9 &
53.5 \\
PartSAM & 37.2 & 42.6 & 42.2 & 40.5 & 69.5 \\
P3-SAM & 40.3 & 41.4 & 40.3 & 38.7 & 59.9 \\
SegviGen & 34.2 & 43.8 & 37.2 & 35.4 & 50.6 \\
\midrule
\textbf{Ours} & \textbf{56.7} & \textbf{86.1} & \textbf{74.2} &
\textbf{68.3} & \textbf{78.8} \\
\bottomrule
\end{tabular}
\end{table}

\begin{table*}[t]
\centering
\caption{
Quantitative results on semantic full-shape segmentation.
We report semantic-aware mIoU (SA-mIoU) together with mask-only mIoU.
}
\label{tab:semantic_full_results}
\small
\setlength{\tabcolsep}{7pt}
\begin{tabular}{lcccccc}
\toprule
\multirow{2}{*}{Method} & \multicolumn{2}{c}{PartNeXt} & \multicolumn{2}{c}{3DCoMPaT200 Coarse} & \multicolumn{2}{c}{3DCoMPaT200 Fine} \\
\cmidrule(lr){2-3} \cmidrule(lr){4-5} \cmidrule(lr){6-7}
 & Mask mIoU & SA-mIoU & Mask mIoU & SA-mIoU & Mask mIoU & SA-mIoU \\
\midrule
PartSAM+MLLM & 37.2 & 13.8 & 42.6 & 25.1 & 42.2 & 20.8 \\
P3-SAM+MLLM & 40.3 & 18.3 & 41.4 & 23.5 & 40.3 & 16.7 \\
\midrule
\textbf{Ours} & \textbf{56.7} & \textbf{43.2} & \textbf{86.1} & \textbf{71.9} & \textbf{74.2} & \textbf{63.1} \\
\bottomrule
\end{tabular}
\end{table*}

\subsubsection{Semantic Full-Shape Segmentation}
Beyond category-agnostic mask quality, this setting further evaluates whether
the predicted full-shape parts are assigned correct semantic labels.
Existing open-world part segmentation methods do not directly support this
setting, since they either produce category-agnostic masks or require queried
part names.
We therefore construct two semantic baselines by pairing category-agnostic
full-shape segmentation methods with an MLLM-based naming stage.
Specifically, we take the masks generated by PartSAM and P3-SAM, render each
predicted mask as a highlighted region on the input shape, and ask an MLLM~\cite{qwen3vl} to
predict the semantic part name of the highlighted region.
This yields PartSAM+MLLM and P3-SAM+MLLM, which add semantic labels to
category-agnostic full-shape segmentations.
To evaluate both geometry and semantics, we report semantic-aware mIoU
(SA-mIoU).
Let $\mathcal{G}$ and $\hat{\mathcal{G}}$ denote the ground-truth and predicted
part sets, where each part is represented by a mask and a semantic label.
We define SA-mIoU as
\begin{equation}
\mathrm{SA\text{-}mIoU} =
\frac{1}{|\mathcal{G}|}
\sum_{(M,y)\in\mathcal{G}}
\max_{(\hat{M},\hat{y})\in\hat{\mathcal{G}}}
\mathbb{I}_{\mathrm{sem}}(y,\hat{y})\,
\mathrm{IoU}(M,\hat{M}).
\end{equation}
Because part names may be ambiguous, we use an LLM-based matcher to decide
the semantic match indicator $\mathbb{I}_{\mathrm{sem}}$.
The matcher is given the object context and the two part names, and determines
whether the predicted name should be considered a correct match to the
ground-truth name.
Specifically, we use GPT-5.5 through the OpenAI API as the semantic matcher,
which is independent of the Qwen3-VL backbone used by PartLLM. We use the
following prompt:
\begin{quote}
\footnotesize
For a 3D object of category ``\{object\_category\}'', determine whether the
predicted part name ``\{pred\_name\}'' and the ground-truth part name
``\{gt\_name\}'' are semantically equivalent part names. Answer only ``yes''
or ``no''.
\end{quote}
This metric penalizes both geometric mismatch and semantic mislabeling: a mask
receives credit only when it overlaps the target part and is assigned a
compatible semantic name.

\paragraph{Results.}
Table~\ref{tab:semantic_full_results} reports the results.
PartLLM obtains the highest mask mIoU and SA-mIoU in all three settings.
More importantly, its performance drops much less when moving from mask-only
mIoU to SA-mIoU.
The two-stage baselines drop by 17.5--23.6 points after semantic matching,
indicating that their masks are often not aligned with semantically meaningful
part identities even when they overlap reasonable geometric regions.
In contrast, PartLLM drops by only 11.1--14.2 points, suggesting that its
predicted regions are already tied to the part names generated by the model.
This supports our formulation: full-shape part segmentation should predict
masks and semantics together, so that the output decomposition is not only
geometrically accurate but also semantically meaningful.

\begin{figure*}[!t]
  \centering
  \includegraphics[width=0.9\textwidth]{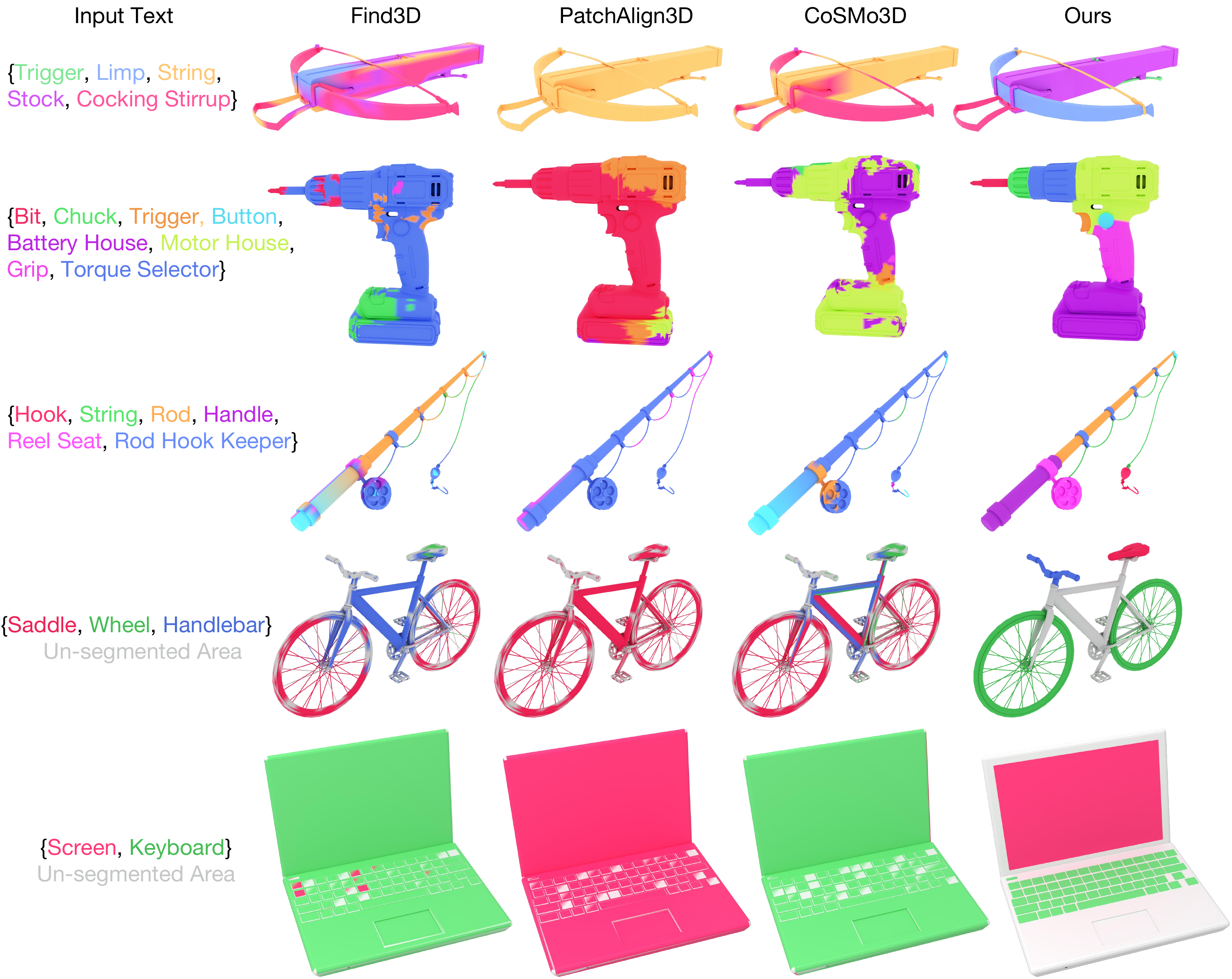}
  \caption{
  Qualitative comparison with baseline methods~\cite{ma2024find,hadgi2026patchalign3d,jin2026cosmo3d} on text-guided part segmentation.
  }
  \Description{Text-guided part segmentation comparison among Find3D,
  PatchAlign3D, CoSMo3D, and PartLLM on a strap, drill, fishing rod, bicycle,
  and laptop. Input part names are listed at left, and colors show the
  predicted regions.}
  \label{fig:text_guided_comp}
\end{figure*}

\subsection{Evaluation on Text-Guided Part Segmentation}
\label{sec:exp-text_guided}

This task evaluates whether a model can segment the requested parts in a 3D
shape given queried part names.
For this task, the input to PartLLM is a set of queried part names, and the
model outputs masks for the requested parts.
We compare our method with three state-of-the-art methods:
Find3D~\cite{ma2024find}, PatchAlign3D~\cite{hadgi2026patchalign3d}, and CoSMo3D~\cite{jin2026cosmo3d}.
For quantitative evaluation, we feed all ground-truth part names to the model
at once and compute metrics from the corresponding output masks.
We report category-averaged mIoU as the evaluation metric.

\begin{table}[!b]
\centering
\caption{
Quantitative results on text-guided part segmentation.
}
\label{tab:text_guided_results}
\small
\setlength{\tabcolsep}{5pt}
\begin{tabular}{lcccc}
\toprule
\multirow{2}{*}{Method} & \multirow{2}{*}{PartNet-E} & \multirow{2}{*}{PartNeXt} & \multicolumn{2}{c}{3DCoMPaT200} \\
\cmidrule(lr){4-5}
 &  &  & Coarse & Fine \\
\midrule
FIND3D & 16.7 & 28.8 & 31.4 & 10.4 \\
PatchAlign3D & 41.4 & 29.3 & 32.8 & 10.3 \\
CoSMo3D & 17.6 & 37.9 & 48.8 & 26.1 \\
\midrule
\textbf{Ours} & \textbf{50.1} & \textbf{78.7} & \textbf{86.1} & \textbf{80.0} \\
\bottomrule
\end{tabular}
\end{table}

\paragraph{Results.}
Table~\ref{tab:text_guided_results} summarizes the results.
PartLLM outperforms all baselines on every benchmark by a large margin.
The gains are particularly striking on PartNeXt (+40.8 over CoSMo3D) and 3DCoMPaT200 (+37.3 and +53.9 at coarse and fine granularities).
Fig.~\ref{fig:text_guided_comp} shows qualitative results.
The first three rows query all parts of the object simultaneously, and the last two rows query only a subset.
Baselines often produce fragmented or incomplete masks, while PartLLM produces consistently accurate masks with sharp boundaries in both cases.
Non-queried regions are cleanly absorbed by the \texttt{<|BG|>} query (shown
in gray), demonstrating that the proposed decoding strategy produces mutually
exclusive masks with precise boundaries regardless of how many parts are
queried.
We attribute these advantages to the formulation of PartLLM.
Existing methods treat each queried part name as an independent
text-to-geometry matching problem, making them sensitive to ambiguous or
compositional names whose interpretation depends on object context and on the
other queried parts.
PartLLM instead predicts all requested parts within a single semantic
decomposition, so each mask is inferred jointly with the others and with the
overall shape.
This leads to more accurate part disambiguation, cleaner exclusion of
non-queried regions, and more mutually consistent masks.

\begin{table*}[t]
\centering
\caption{
Quantitative results on interactive part segmentation.
IoU@$k$ denotes mean IoU after $k$ interaction rounds.
}
\label{tab:point_results}
\setlength{\tabcolsep}{3pt}
\begin{tabular}{l cccc cccc cccc cccc}
\toprule
 & \multicolumn{4}{c}{\multirow{2}{*}{PartNeXt}} & \multicolumn{8}{c}{3DCoMPaT200} & \multicolumn{4}{c}{\multirow{2}{*}{HY3D-Bench}} \\
\cmidrule(lr){6-13}
 & \multicolumn{4}{c}{} & \multicolumn{4}{c}{Coarse} & \multicolumn{4}{c}{Fine} & \multicolumn{4}{c}{} \\
\cmidrule(lr){2-5} \cmidrule(lr){6-9} \cmidrule(lr){10-13} \cmidrule(lr){14-17}
Method & @1 & @3 & @5 & @7 & @1 & @3 & @5 & @7 & @1 & @3 & @5 & @7 & @1 & @3 & @5 & @7 \\
\midrule
Point-SAM & 36.6 & 49.0 & 55.0 & 60.3 & 45.6 & 65.9 & 71.1 & 74.3 & 37.3 & 46.9 & 49.4 & 51.6 & 20.7 & 26.2 & 29.7 & 31.1 \\
P3-SAM & 43.8 & - & - & - & 47.7 & - & - & - & 49.9 & - & - & - & 31.8 & - & - & - \\
S2AM3D & 40.8 & - & - & - & 62.6 & - & - & - & 43.3 & - & - & - & 39.8 & - & - & - \\
PartSAM & 45.9 & 61.3 & 65.0 & 69.6 & 51.3 & 76.8 & 79.6 & 81.8 & 50.6 & 66.5 & 71.8 & 73.5 & 34.6 & 45.1 & 49.6 & 50.4 \\
SegviGen & 49.8 & 66.8 & 70.0 & 71.7 & 67.9 & 75.5 & 82.0 & 83.6 & 45.3 & 52.8 & 62.7 & 77.6 & 43.7 & 52.9 & 56.2 & 58.6 \\
\midrule
\textbf{Ours} & \textbf{58.9} & \textbf{73.4} & \textbf{79.1} & \textbf{83.5} & \textbf{70.6} & \textbf{84.1} & \textbf{88.9} & \textbf{91.4} & \textbf{60.7} & \textbf{75.3} & \textbf{78.8} & \textbf{81.2} & \textbf{51.2} & \textbf{67.7} & \textbf{72.4} & \textbf{77.1} \\
\bottomrule
\end{tabular}
\end{table*}

\begin{figure*}[!t]
  \centering
  \includegraphics[width=0.95\textwidth]{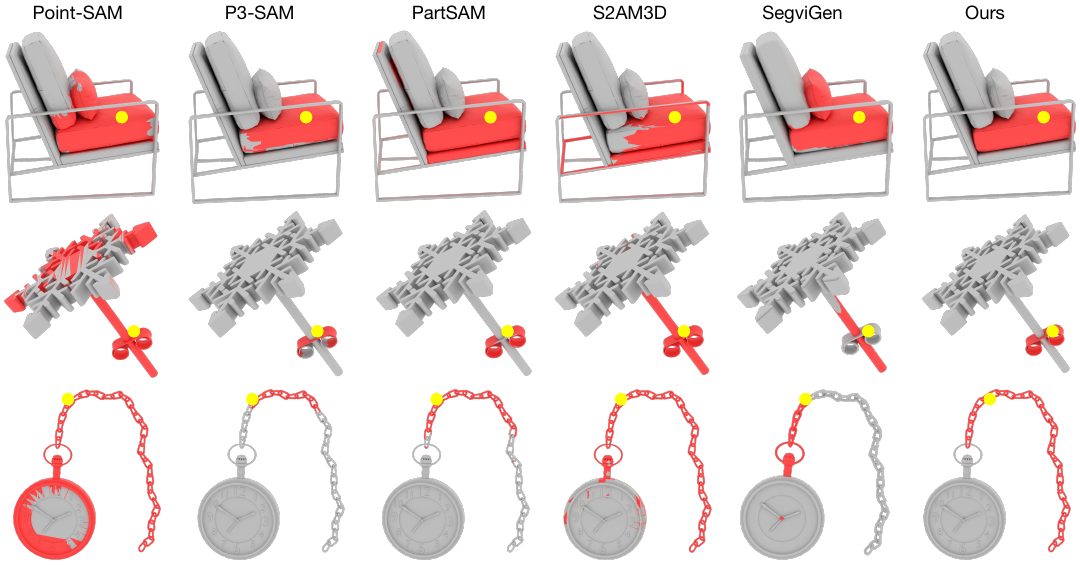}
  \caption{
  Qualitative comparison with baseline methods~\cite{zhou2025pointsam,p3sam,zhu2026partsam,s2am3d,li2026segvigen} on one-click interactive part segmentation.
  Yellow dots denote prompt points, and red regions denote predicted masks.
  }
  \Description{One-click interactive segmentation comparison among Point-SAM,
  P3-SAM, PartSAM, S2AM3D, SegViGen, and PartLLM on three objects. Yellow dots
  mark the input clicks and red regions show the predicted masks.}
  \label{fig:interactive_comp}
\end{figure*}

\subsection{Evaluation on Interactive Part Segmentation}
\label{sec:exp-point}

Interactive part segmentation evaluates whether a model can segment the target
part indicated by sparse clicks.
Following the experimental protocol of previous works~\cite{zhou2025pointsam,zhu2026partsam}, we
evaluate with multi-round interactions and report mIoU after each round.
For each ground-truth mask, the first point is sampled from its central
region; subsequent points are iteratively selected from the error
regions between the predicted mask and the ground truth.
For the first click, the input prompt is ``Please segment the part at \texttt{(x, y, z)} in \texttt{<point\_cloud>},'' and the model outputs the corresponding part mask; 
for later clicks, the current mask is encoded as point colors in \texttt{<point\_cloud>}, and the prompt provides an include/exclude corrective point at \texttt{(x, y, z)} to produce a refined mask.
We compare with Point-SAM~\cite{zhou2025pointsam},
P3-SAM~\cite{p3sam}, S2AM3D~\cite{s2am3d},
PartSAM~\cite{zhu2026partsam}, and SegViGen~\cite{li2026segvigen}.
Point-SAM, PartSAM, and SegViGen are evaluated with the multi-round protocol;
for the remaining baselines, we report the available one-click results.

\paragraph{Results.}
Table~\ref{tab:point_results} reports the quantitative results.
PartLLM achieves the best mIoU on every reported dataset and at every interaction round.
With only one click, it improves over the strongest baseline by up to 10.1 mIoU points;
after seven clicks, the margin reaches up to 18.5 mIoU among methods with
multi-round outputs.
These gains show that PartLLM can infer the complete target part from a sparse
prompt rather than treating the click as only a local geometric seed, and can
use corrective prompts to refine the mask while preserving the semantic extent
of the target part.
Fig.~\ref{fig:interactive_comp} shows qualitative one-click comparisons.
Given the same prompt point, baselines often either capture only a local region
around the click or leak into adjacent structures with similar geometry.
PartLLM instead recovers complete parts with cleaner boundaries, including thin
or spatially extended structures where local geometric cues alone are
ambiguous.
These results suggest that the click is more effective when interpreted as part
of a semantic decomposition rather than as an isolated geometric seed.
This helps PartLLM separate the target part from neighboring regions with
similar local geometry.

\subsection{Evaluation on Unseen Real-World Scans}
\label{sec:exp-real-scans}

\begin{figure}[t]
  \centering
  \includegraphics[width=\columnwidth]{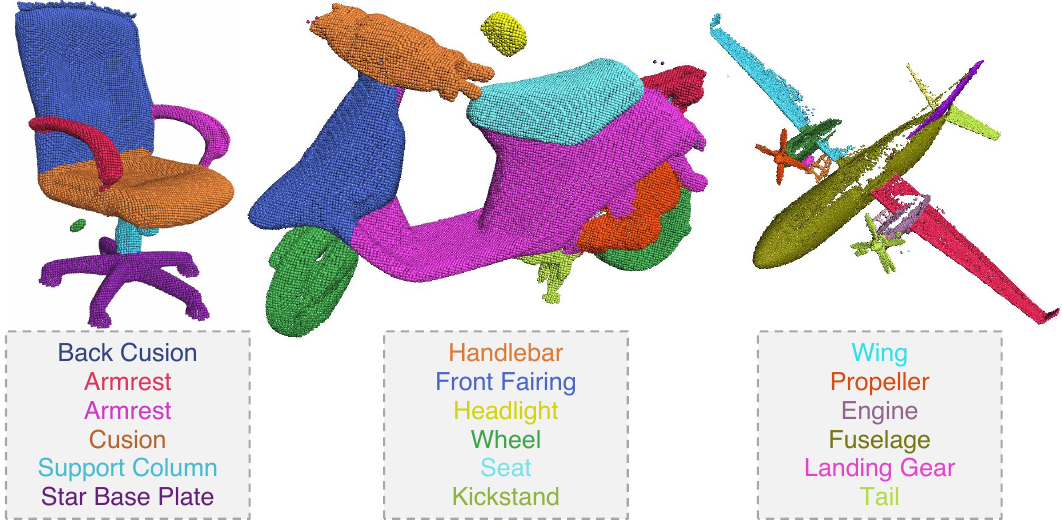}
  \caption{
  Results on real-world scans with noise and partial structure.
  Despite missing surfaces and capture artifacts, PartLLM still produces usable
  semantic part decompositions.
  }
  \Description{PartLLM results on noisy and partial real-world scans.}
  \label{fig:raw_scan}
\end{figure}

To evaluate generalization beyond synthetic assets, we conduct
additional experiments on two benchmarks containing unseen real-world scans.
FAUST contains human scans with part annotations adopted from
SATR~\cite{abdelreheem2023satr}, while AKBSeg contains selected articulated-part
annotations on scanned AKB-48 objects~\cite{liu2022akb}. We evaluate category-agnostic full-shape
segmentation on FAUST against PartSAM~\cite{zhu2026partsam},
P3-SAM~\cite{p3sam}, and SegviGen~\cite{li2026segvigen}. We further evaluate
text-guided segmentation on both benchmarks against
PartSLIP~\cite{liu2023partslip}, ZeroPS~\cite{xue2025zerops},
SATR~\cite{abdelreheem2023satr}, and
PatchAlign3D~\cite{hadgi2026patchalign3d}.
As shown in Tables~\ref{tab:real_scan_full_results}
and~\ref{tab:real_scan_text_results}, PartLLM achieves the best performance in
every setting. Fig.~\ref{fig:raw_scan} further presents results on
scans with sensor noise, partial geometry, and surface irregularities.
PartLLM still produces coherent segmentation results despite these capture
artifacts.

\begin{table}[t]
\centering
\caption{
Category-agnostic full-shape segmentation on FAUST.
}
\label{tab:real_scan_full_results}
\setlength{\tabcolsep}{5pt}
\begin{tabular}{lcccc}
\toprule
Dataset & PartSAM & P3-SAM & SegviGen & Ours \\
\midrule
FAUST & 55.3 & 35.0 & 37.7 & \textbf{72.2} \\
\bottomrule
\end{tabular}
\end{table}

\begin{table}[t]
\centering
\caption{
Text-guided segmentation on unseen real-world scans.
}
\label{tab:real_scan_text_results}
\setlength{\tabcolsep}{8pt}
\begin{tabular}{lcc}
\toprule
Method & FAUST & AKBSeg \\
\midrule
PartSLIP & 48.6 & 22.6 \\
ZeroPS & 49.5 & 33.5 \\
SATR & 79.8 & 29.0 \\
PatchAlign3D & 66.5 & 35.6 \\
\textbf{Ours} & \textbf{87.6} & \textbf{54.8} \\
\bottomrule
\end{tabular}
\end{table}

\subsection{Ablation and Analysis}
\label{sec:exp-ablation}

\subsubsection{Method Ablations}

\begin{table*}[t]
\centering
\caption{
Ablation study on method design choices. All entries use mIoU
except Semantic, which uses SA-mIoU.
}
\label{tab:method_ablation}
\small
\setlength{\tabcolsep}{5pt}
\begin{tabular}{@{}lcccccc@{}}
\toprule
\multirow{2}{*}{Variant} & \multicolumn{4}{c}{3DCoMPaT200-Fine} &
\multirow{2}{*}{PartObjaverse-Tiny} &
\multirow{2}{*}{PartNet-E} \\
\cmidrule(lr){2-5}
 & Text & Interactive & Full & Semantic & & \\
\midrule
w/ SAM-style decoder & 76.2 & 57.5 & 62.9 & 54.6 &
67.0 & 47.7 \\
w/o coarse 3D location & 55.4 & 46.3 & 57.6 & 38.1 &
61.2 & 34.8 \\
w/o semantic label & 62.0 & 53.8 & 61.3 & -- &
65.0 & 38.8 \\
\midrule
Full PartLLM & \textbf{80.0} & \textbf{60.7} & \textbf{74.2} &
\textbf{63.1} & \textbf{78.8} &
\textbf{50.1} \\
\bottomrule
\end{tabular}
\end{table*}

\begin{figure}[!htbp]
  \centering
  \includegraphics[width=\columnwidth]{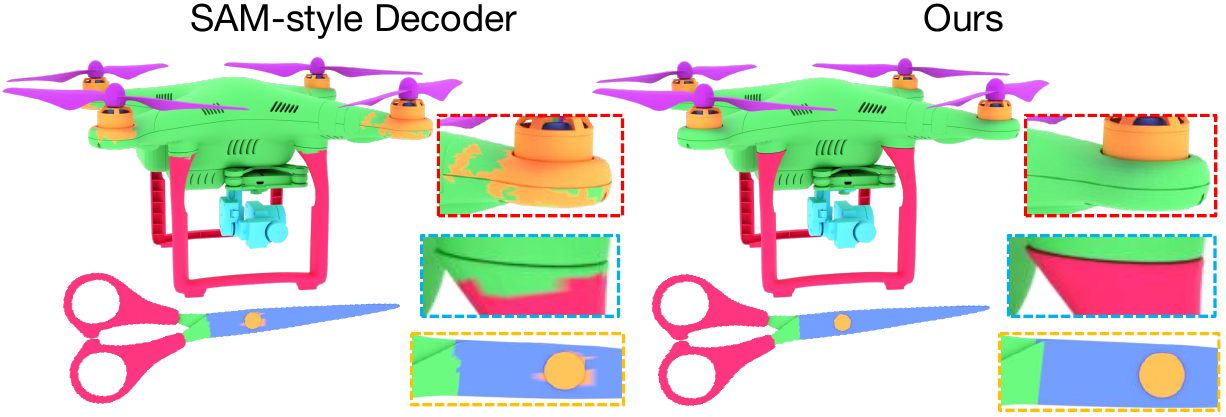}
  \caption{
  Qualitative ablation of the decoding strategy.
  Compared with the SAM-style decoder that predicts masks independently, our decoder produces cleaner mutually exclusive regions with
  sharper boundaries.
  }
  \Description{Side-by-side qualitative comparison of a SAM-style decoder and
  the PartLLM decoder on a drone and scissors. Enlarged regions show that the
  PartLLM decoder produces cleaner boundaries between adjacent parts.}
  \label{fig:decoder_ablation}
\end{figure}

Table~\ref{tab:method_ablation} evaluates the main design choices
of PartLLM. The last two columns further report full-shape mask mIoU on
PartObjaverse-Tiny and text-guided mIoU on PartNet-E. We test the
decomposition-aware decoder by replacing it with an
independent binary mask predictor and removing softmax competition among the
part and background queries.
This variant examines whether decomposition-aware point assignment provides benefits beyond SAM-style per-query mask prediction.
As shown in the table, while this modification causes only marginal performance drops on text-guided and interactive segmentation, it substantially reduces full-shape mask mIoU and SA-mIoU.
The result suggests that independent binary decoding can still handle a small number of prompted targets, but it is less suitable for complete shape decomposition, where all generated parts must compete for points under a globally consistent assignment.
Fig.~\ref{fig:decoder_ablation} further illustrates this effect
qualitatively: independent decoding tends to produce ambiguous boundaries near
adjacent parts, whereas the decomposition-aware decoder separates parts with
sharper boundaries.

We next ablate the information encoded in each part hypothesis.
As shown in the table, removing the coarse 3D location leads to the largest degradation in most settings, indicating that semantic labels alone are insufficient for localizing parts within complex shapes.
Furthermore, removing the semantic label also consistently harms all evaluable metrics, especially text-guided segmentation,
because the model no longer receives direct supervision to align part names with their corresponding geometry.
This degradation is naturally less pronounced in interactive segmentation, where the input click intrinsically serves as a strong spatial anchor.
Overall, these results indicate that PartLLM benefits from generating
part hypotheses that are both semantically grounded and spatially localized,
and from resolving them through a decomposition-aware decoder.

\subsubsection{Robustness to Rotation}

To evaluate sensitivity to object orientation, we apply random
rotations around all three axes at test time and compare performance under
random rotations with that under the canonical orientation. As shown in Table~\ref{tab:rotation_robustness}, random rotations
result in negligible performance degradation in both settings. PartLLM also
maintains clear margins over PartSAM after rotation, indicating that its
performance is not tied to a canonical object orientation.

\subsubsection{Robustness to Probabilistic Sampling}

PartLLM uses probabilistic autoregressive sampling, so different random seeds
may produce different part-hypothesis sequences. In the main experiments, we use fixed sampling
parameters and a fixed random seed for all reported comparisons. To assess
sensitivity to the sampling seed, we repeat full-shape inference on
PartObjaverse-Tiny using ten random seeds. PartLLM obtains
$78.5 \pm 2.2$ mIoU (mean $\pm$ standard deviation), close to the reported
result, indicating stable performance across random seeds.

\subsubsection{Robustness to Prompt Variations}

We first examine whether text-guided segmentation requires the complete set of
part names as input. Table~\ref{tab:prompt_variations} compares the standard
all-part setting with a single-part setting in which only one target name is
queried. Performance remains comparable across all benchmarks, showing that
PartLLM can segment individual queried parts without requiring the full part
vocabulary. We further test prompt paraphrases for
full-shape segmentation on PartObjaverse-Tiny. Replacing ``main parts'' with
``coarse structures'' and ``detailed parts'' with ``fine-grained components''
yields 76.2 mIoU, compared with 78.8 using the default templates. This small
difference indicates that the prompt templates are default
formulations rather than strict input formats. Overall, these results
demonstrate that PartLLM is robust to prompt variations and does not rely on a
specific prompt formulation.

\subsubsection{Data Composition Ablations}

Table~\ref{tab:data_ablation} studies the effect of multi-task data integration in our unified part segmentation framework.
Under the same architecture and training objective, we progressively add
training data from the four supervision sources.
This ablation evaluates whether each data source only benefits its matched setting or transfers to other part segmentation tasks.

Starting from full-shape segmentation data, we sequentially add text-guided data, interactive data, and category-agnostic geometry data.
As shown in the table, adding text-guided data not only enables text-guided evaluation but also improves full-shape mask mIoU from 56.2 to 62.7 and SA-mIoU from 48.3 to 58.5, indicating that explicit part-name supervision promotes more semantically meaningful decompositions.
Adding interactive data further improves both text-guided and full-shape segmentation.
This suggests that point-conditioned supervision strengthens part localization and transfers to non-interactive settings.
Finally, adding category-agnostic training data improves all metrics, with especially large gains on full-shape segmentation and interactive segmentation.
Although these data do not provide semantic part names, they supply additional geometric decomposition supervision, improving the model's boundary and shape priors.
These trends support the motivation of our unified formulation:
heterogeneous task data can reinforce a shared representation of 3D parts
rather than merely improving performance on their corresponding tasks.

\subsubsection{Scaling Analysis}

\begin{figure}[t]
  \centering
  \includegraphics[width=\columnwidth]{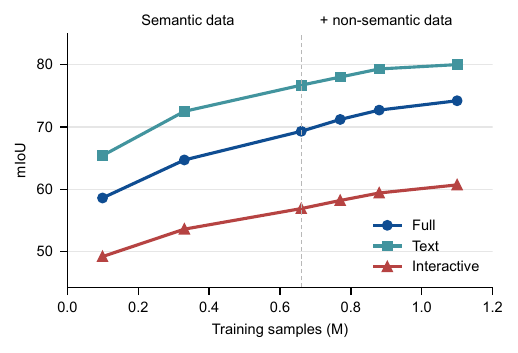}
  \caption{
  Scaling curve on the fine-grained split of 3DCoMPaT200.
  }
  \label{fig:scaling_analysis}
  \Description{Line plot of Full, Text, and Interactive scores as training data scale increases.}
\end{figure}

Fig.~\ref{fig:scaling_analysis} evaluates how PartLLM scales with the volume of unified training data.
The scaling curve shows consistent gains across all three evaluation settings.
Notably, scaling up solely the semantic part data yields steady improvements across all tasks, indicating that our unified formulation effectively capitalizes on both task diversity and expanded semantic supervision.
Furthermore, holding the semantic data constant and introducing category-agnostic part decomposition data~\cite{hunyuan3d2026hy3dbench} yields further performance boosts across the board.
These gains are particularly pronounced in full-shape and interactive segmentation, where the influx of geometric decomposition examples directly refines boundary delineation and part-extent localization.
Interestingly, text-guided segmentation also exhibits marginal improvements, suggesting that enhanced geometric part priors positively transfer to semantic understanding, even in the absence of explicit semantic labels in the added data.
This result supports our data scaling argument: heterogeneous part data can be absorbed by the same
formulation and converted into broadly useful 3D part understanding.

\begin{table}[!b]
\centering
\caption{
Robustness to random 3D rotations. We report full-shape mIoU on
PartObjaverse-Tiny and one-click mIoU on HY3D-Bench.
}
\label{tab:rotation_robustness}
\setlength{\tabcolsep}{5pt}
\begin{tabular}{llcc}
\toprule
Task & Setting & PartSAM & Ours \\
\midrule
\multirow{2}{*}{Full} & Canonical & 69.5 & \textbf{78.8} \\
 & Rotated & 67.9 & \textbf{78.3} \\
\midrule
\multirow{2}{*}{Interactive} & Canonical & 34.6 & \textbf{51.2} \\
 & Rotated & 35.3 & \textbf{50.7} \\
\bottomrule
\end{tabular}
\end{table}

\begin{table}[t]
\centering
\caption{
Text-guided mIoU with all-part and single-part queries.
}
\label{tab:prompt_variations}
\setlength{\tabcolsep}{4pt}
\begin{tabular}{lcc}
\toprule
Dataset & All-part & Single-part \\
\midrule
PartNet-E & 50.1 & \textbf{53.3} \\
PartNeXt & \textbf{78.7} & 77.0 \\
3DCoMPaT200-Coarse & 86.1 & \textbf{87.4} \\
3DCoMPaT200-Fine & \textbf{80.0} & 78.2 \\
\bottomrule
\end{tabular}
\end{table}

\begin{table}[t]
\centering
\caption{
Ablation study on training data
on 3DCoMPaT200-Fine.
}
\label{tab:data_ablation}
\setlength{\tabcolsep}{1.5pt}
\begin{tabular}{@{}cccccccc@{}}
\toprule
\multicolumn{4}{c}{Training data} & \multicolumn{4}{c}{Evaluation} \\
\cmidrule(lr){1-4} \cmidrule(l){5-8}
Full & Text & Interactive & Geometry & Full & Semantic & Text & Interactive \\
\midrule
\cmark & & & & 56.2 & 48.3 & - & - \\
\cmark & \cmark & & & 62.7 & 58.5 & 72.0 & - \\
\cmark & \cmark & \cmark & & 67.2 & 61.8 & 77.5 & 56.5 \\
\cmark & \cmark & \cmark & \cmark & \textbf{74.2} & \textbf{63.1} & \textbf{80.0} & \textbf{60.7} \\
\bottomrule
\end{tabular}
\end{table}

\subsubsection{Efficiency Analysis}

We report the running time comparison in Table~\ref{tab:runtime_comparison}.
All runtime measurements in Table~\ref{tab:runtime_comparison} are obtained on
a single NVIDIA H20 GPU.
While PartLLM relies on an autoregressive generation paradigm,
modern LLM inference is highly optimized and benefits from inference
acceleration frameworks~\cite{vllm}.
In practice, for typical 3D objects with a moderate number of
semantic parts, PartLLM achieves inference efficiency competitive with or even
superior to that of prior baselines, while attaining substantially higher
full-shape segmentation accuracy.
Although generation latency naturally scales with the requested
part count, our framework maintains a favorable trade-off between
computational efficiency and semantic segmentation quality.

\begin{table}[t]
\centering
\caption{
Runtime comparison on the full-shape segmentation task
on 3DCoMPaT200-Fine.
$P$ denotes the number of parts.
}
\label{tab:runtime_comparison}
\setlength{\tabcolsep}{1.5pt}
\begin{tabular}{lccccc}
\toprule
\multirow{2}{*}{Method} & \multicolumn{4}{c}{Runtime (s)} & \multirow{2}{*}{mIoU} \\
\cmidrule(lr){2-5}
 & $P{=}5$ & $P{=}10$ & $P{=}20$ & $P{=}30$ &  \\
\midrule
PartField~\cite{liu2025partfield} & \multicolumn{4}{c}{$\sim$10} & 35.0 \\
PartSAM~\cite{zhu2026partsam} & \multicolumn{4}{c}{$\sim$12} & 42.2 \\
P3-SAM~\cite{p3sam} & \multicolumn{4}{c}{$\sim$10} & 40.3 \\
\midrule
Ours & 4.8 & 7.1 & 11.7 & 16.3 & 74.2 \\
\bottomrule
\end{tabular}
\end{table}

\subsection{Application}
\label{sec:exp-application}

\begin{figure}[t]
  \centering
  \includegraphics[width=\columnwidth]{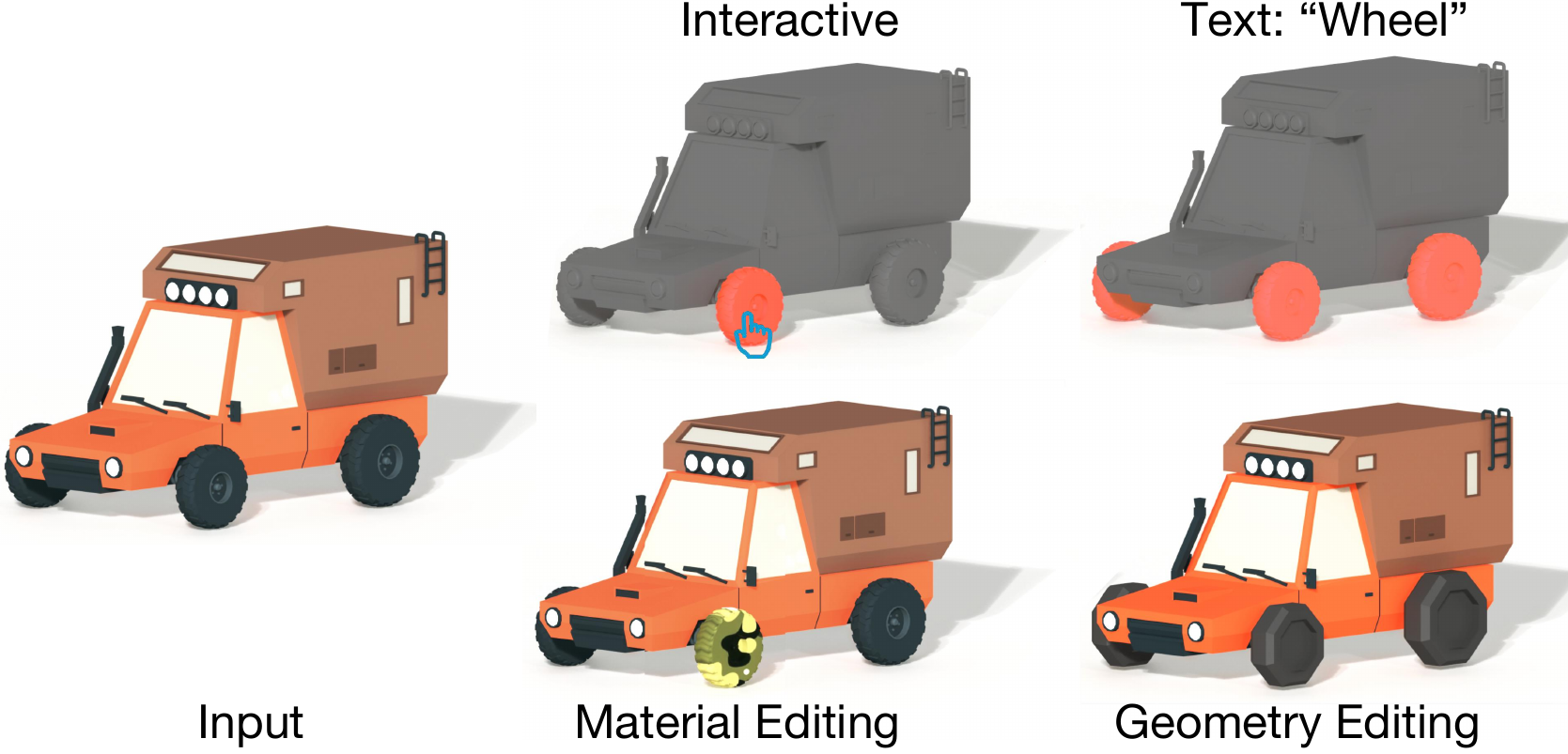}
  \caption{
  Editing applications enabled by PartLLM.
  A target part can be selected either interactively or through a semantic text
  query such as \texttt{Wheel}, and the resulting mask can serve as a direct
  handle for downstream material and geometry editing.
  }
  \Description{Part-aware editing application figure showing part selection by interactive prompting or text query, followed by material editing and geometry editing on the selected part.}
  \label{fig:application_editing}
\end{figure}

PartLLM exposes its unified 3D part understanding through a semantic
decomposition interface, enabling practical use beyond standard benchmark
settings. PartLLM generalizes to real-world scans, as demonstrated in
Sec.~\ref{sec:exp-real-scans}, and supports part-aware editing, where the
predicted masks serve as controllable handles for modifying 3D assets.
Fig.~\ref{fig:application_editing} shows two editing workflows
enabled by this interface.
A target part can be localized either from an interactive click or from a
semantic text query, after which the predicted mask is reused as a controllable
editing region.
This makes the output immediately actionable: the same part handle can support
material replacement as well as structural edits that
modify the part geometry.

\subsection{Limitations and Future Work}
\label{sec:exp-limitations}

\begin{figure}[H]
  \centering
  \includegraphics[width=\columnwidth]{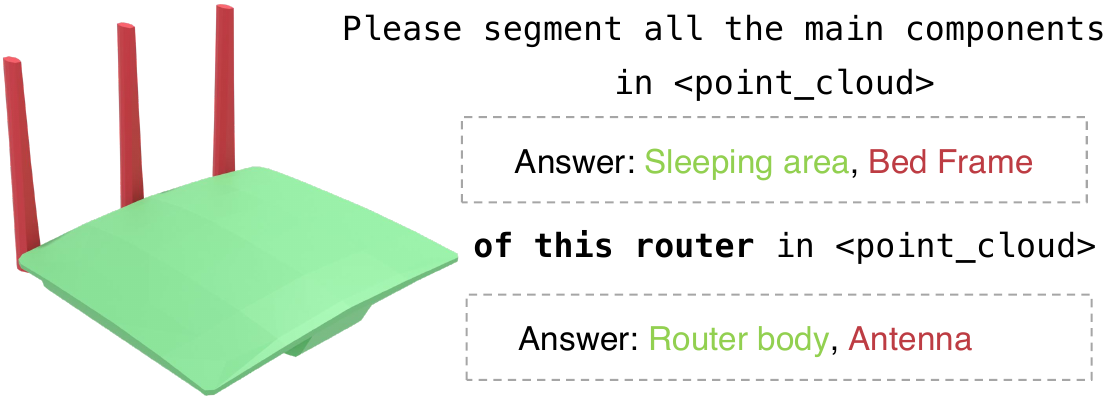}
  \caption{
  Representative failure case of category confusion.
  When the object-level prior is incorrect, PartLLM may produce a semantically coherent but wrong decomposition. Adding the object category to the prompt can correct the prediction.
  }
  \Description{Failure case figure showing a router that is incorrectly decomposed with bed-related part names, followed by a corrected prediction when the object category is provided in the prompt.}
  \label{fig:failure_cases}
\end{figure}

As shown in Fig.~\ref{fig:failure_cases}, a representative failure mode arises
when the object category is ambiguous from geometry alone.
In such cases, PartLLM may produce a decomposition that is internally coherent
but organized under the wrong semantic prior.
For instance, misclassifying a router as a bed causes the model to hallucinate
bed-associated sub-structures over the router's topology.
Providing the object category as an additional prompt cue can disambiguate the
prediction and recover the correct part decomposition, including the router
body and antennas.
This suggests that stronger object-level context is important for precise
semantic part understanding.
Future work will explore integrating additional modalities, such as rendered
images or multi-view visual features, to provide stronger category evidence.
Furthermore, incorporating chain-of-thought (CoT) reasoning could stabilize high-level semantic decisions prior to executing dense mask predictions.

\section{Conclusion}

We presented PartLLM, a unified framework that reformulates 3D part
segmentation as autoregressive semantic decomposition and realizes this
formulation through a multimodal large language model.
By using language as the interface for part decomposition and mask
prediction, PartLLM supports open-vocabulary and granularity-controllable part
understanding within a single generate-and-ground pipeline.
Experiments show that this formulation does more than integrate heterogeneous
supervision: it turns task and data diversity into a new scaling axis for 3D
part understanding, consistently outperforming strong task-specific baselines
as training data expands.
More broadly, our results suggest a path beyond task-specific 3D segmentation
models toward a foundation model paradigm for part-level understanding, where
parts serve as a general semantic and operational interface for open-world 3D
assets.

\begin{acks}
This work was supported by the National Natural Science Foundation of China
(No. T2322012, No. 62572240, No. 62172218).
It was also supported by the Macao Science and Technology Development Fund (FDCT) (0119/2025/ITP2).
\end{acks}

\bibliographystyle{ACM-Reference-Format}
\bibliography{reference}

\begin{thebibliography}{59}

\ifx \showCODEN    \undefined \def \showCODEN     #1{\unskip}     \fi
\ifx \showISBNx    \undefined \def \showISBNx     #1{\unskip}     \fi
\ifx \showISBNxiii \undefined \def \showISBNxiii  #1{\unskip}     \fi
\ifx \showISSN     \undefined \def \showISSN      #1{\unskip}     \fi
\ifx \showLCCN     \undefined \def \showLCCN      #1{\unskip}     \fi
\ifx \shownote     \undefined \def \shownote      #1{#1}          \fi
\ifx \showarticletitle \undefined \def \showarticletitle #1{#1}   \fi
\ifx \showURL      \undefined \def \showURL       {\relax}        \fi
\providecommand\bibfield[2]{#2}
\providecommand\bibinfo[2]{#2}
\providecommand\natexlab[1]{#1}
\providecommand\showeprint[2][]{arXiv:#2}

\bibitem[Abdelreheem et~al\mbox{.}(2023)]%
        {abdelreheem2023satr}
\bibfield{author}{\bibinfo{person}{Ahmed Abdelreheem}, \bibinfo{person}{Ivan
  Skorokhodov}, \bibinfo{person}{Maks Ovsjanikov}, {and} \bibinfo{person}{Peter
  Wonka}.} \bibinfo{year}{2023}\natexlab{}.
\newblock \showarticletitle{Satr: Zero-shot semantic segmentation of 3d
  shapes}. In \bibinfo{booktitle}{\emph{Proceedings of the IEEE/CVF
  International Conference on Computer Vision}}. \bibinfo{pages}{15166--15179}.
\newblock


\bibitem[Ahmed et~al\mbox{.}(2024)]%
        {3dcompat200}
\bibfield{author}{\bibinfo{person}{Mahmoud Ahmed}, \bibinfo{person}{Xiang Li},
  \bibinfo{person}{Arpit Prajapati}, {and} \bibinfo{person}{Mohamed
  Elhoseiny}.} \bibinfo{year}{2024}\natexlab{}.
\newblock \showarticletitle{3{DC}o{MP}aT200: Language Grounded Large-Scale 3D
  Vision Dataset for Compositional Recognition}. In
  \bibinfo{booktitle}{\emph{The Thirty-eight Conference on Neural Information
  Processing Systems Datasets and Benchmarks Track}}.
\newblock
\urldef\tempurl%
\url{https://openreview.net/forum?id=L4yLhMjCOR}
\showURL{%
\tempurl}


\bibitem[Bai et~al\mbox{.}(2023)]%
        {bai2023qwenvl}
\bibfield{author}{\bibinfo{person}{Jinze Bai}, \bibinfo{person}{Shuai Bai},
  \bibinfo{person}{Shusheng Yang}, \bibinfo{person}{Shijie Wang},
  \bibinfo{person}{Sinan Tan}, \bibinfo{person}{Peng Wang},
  \bibinfo{person}{Junyang Lin}, \bibinfo{person}{Chang Zhou}, {and}
  \bibinfo{person}{Jingren Zhou}.} \bibinfo{year}{2023}\natexlab{}.
\newblock \showarticletitle{Qwen-{VL}: A Versatile Vision-Language Model for
  Understanding, Localization, Text Reading, and Beyond}.
\newblock \bibinfo{journal}{\emph{arXiv preprint arXiv:2308.12966}}
  (\bibinfo{year}{2023}).
\newblock


\bibitem[Bai et~al\mbox{.}(2025)]%
        {qwen3vl}
\bibfield{author}{\bibinfo{person}{Shuai Bai}, \bibinfo{person}{Yuxuan Cai},
  \bibinfo{person}{Ruizhe Chen}, \bibinfo{person}{Keqin Chen},
  \bibinfo{person}{Xionghui Chen}, \bibinfo{person}{Zesen Cheng},
  \bibinfo{person}{Lianghao Deng}, \bibinfo{person}{Wei Ding},
  \bibinfo{person}{Chang Gao}, \bibinfo{person}{Chunjiang Ge}, {et~al\mbox{.}}}
  \bibinfo{year}{2025}\natexlab{}.
\newblock \showarticletitle{Qwen3-vl technical report}.
\newblock \bibinfo{journal}{\emph{arXiv preprint arXiv:2511.21631}}
  (\bibinfo{year}{2025}).
\newblock


\bibitem[Chen et~al\mbox{.}(2023)]%
        {chen2023shikra}
\bibfield{author}{\bibinfo{person}{Keqin Chen}, \bibinfo{person}{Zhao Zhang},
  \bibinfo{person}{Weili Zeng}, \bibinfo{person}{Richong Zhang},
  \bibinfo{person}{Feng Zhu}, {and} \bibinfo{person}{Rui Zhao}.}
  \bibinfo{year}{2023}\natexlab{}.
\newblock \showarticletitle{Shikra: Unleashing multimodal llm's referential
  dialogue magic}.
\newblock \bibinfo{journal}{\emph{arXiv preprint arXiv:2306.15195}}
  (\bibinfo{year}{2023}).
\newblock


\bibitem[Chen et~al\mbox{.}(2024)]%
        {chen2023ll3da}
\bibfield{author}{\bibinfo{person}{Sijin Chen}, \bibinfo{person}{Xin Chen},
  \bibinfo{person}{Chi Zhang}, \bibinfo{person}{Mingsheng Li},
  \bibinfo{person}{Gang Yu}, \bibinfo{person}{Hao Fei},
  \bibinfo{person}{Hongyuan Zhu}, \bibinfo{person}{Jiayuan Fan}, {and}
  \bibinfo{person}{Tao Chen}.} \bibinfo{year}{2024}\natexlab{}.
\newblock \showarticletitle{Ll3da: Visual interactive instruction tuning for
  omni-3d understanding reasoning and planning}. In
  \bibinfo{booktitle}{\emph{Proceedings of the IEEE/CVF Conference on Computer
  Vision and Pattern Recognition}}. \bibinfo{pages}{26428--26438}.
\newblock


\bibitem[Cho et~al\mbox{.}(2025)]%
        {cho2024cubellm}
\bibfield{author}{\bibinfo{person}{Jang~Hyun Cho}, \bibinfo{person}{Boris
  Ivanovic}, \bibinfo{person}{Yulong Cao}, \bibinfo{person}{Edward Schmerling},
  \bibinfo{person}{Yue Wang}, \bibinfo{person}{Xinshuo Weng},
  \bibinfo{person}{Boyi Li}, \bibinfo{person}{Yurong You},
  \bibinfo{person}{Philipp Kr{\"a}henb{\"u}hl}, \bibinfo{person}{Yan Wang},
  {et~al\mbox{.}}} \bibinfo{year}{2025}\natexlab{}.
\newblock \showarticletitle{Language-image models with 3d understanding}. In
  \bibinfo{booktitle}{\emph{International Conference on Learning
  Representations}}, Vol.~\bibinfo{volume}{2025}.
  \bibinfo{pages}{36643--36674}.
\newblock


\bibitem[Ding et~al\mbox{.}(2026)]%
        {ding2026fullpart}
\bibfield{author}{\bibinfo{person}{Lihe Ding}, \bibinfo{person}{Shaocong Dong},
  \bibinfo{person}{Yaokun Li}, \bibinfo{person}{Chenjian Gao},
  \bibinfo{person}{Xiao Chen}, \bibinfo{person}{Rui Han},
  \bibinfo{person}{Yihao Kuang}, \bibinfo{person}{Hong Zhang},
  \bibinfo{person}{Bo Huang}, \bibinfo{person}{Zhanpeng Huang},
  \bibinfo{person}{Zibin Wang}, \bibinfo{person}{Dan Xu}, {and}
  \bibinfo{person}{Tianfan Xue}.} \bibinfo{year}{2026}\natexlab{}.
\newblock \showarticletitle{FullPart: Generating each 3D Part at Full
  Resolution}. In \bibinfo{booktitle}{\emph{The Fourteenth International
  Conference on Learning Representations}}.
\newblock
\urldef\tempurl%
\url{https://openreview.net/forum?id=QlRlE7a1p4}
\showURL{%
\tempurl}


\bibitem[Garosi et~al\mbox{.}(2025)]%
        {garosi20253d}
\bibfield{author}{\bibinfo{person}{Marco Garosi}, \bibinfo{person}{Riccardo
  Tedoldi}, \bibinfo{person}{Davide Boscaini}, \bibinfo{person}{Massimiliano
  Mancini}, \bibinfo{person}{Nicu Sebe}, {and} \bibinfo{person}{Fabio Poiesi}.}
  \bibinfo{year}{2025}\natexlab{}.
\newblock \showarticletitle{3d part segmentation via geometric aggregation of
  2d visual features}. In \bibinfo{booktitle}{\emph{2025 IEEE/CVF Winter
  Conference on Applications of Computer Vision}}. \bibinfo{pages}{3257--3267}.
\newblock


\bibitem[Hadgi et~al\mbox{.}(2026)]%
        {hadgi2026patchalign3d}
\bibfield{author}{\bibinfo{person}{Souhail Hadgi}, \bibinfo{person}{Bingchen
  Gong}, \bibinfo{person}{Ramana Sundararaman}, \bibinfo{person}{Emery
  Pierson}, \bibinfo{person}{Lei Li}, \bibinfo{person}{Peter Wonka}, {and}
  \bibinfo{person}{Maks Ovsjanikov}.} \bibinfo{year}{2026}\natexlab{}.
\newblock \showarticletitle{PatchAlign3D: Local Feature Alignment for Dense 3D
  Shape understanding}.
\newblock \bibinfo{journal}{\emph{arXiv preprint arXiv:2601.02457}}
  (\bibinfo{year}{2026}).
\newblock


\bibitem[Hong et~al\mbox{.}(2023)]%
        {hong2023threedllm}
\bibfield{author}{\bibinfo{person}{Yining Hong}, \bibinfo{person}{Haoyu Zhen},
  \bibinfo{person}{Peihao Chen}, \bibinfo{person}{Shuhong Zheng},
  \bibinfo{person}{Yilun Du}, \bibinfo{person}{Zhenfang Chen}, {and}
  \bibinfo{person}{Chuang Gan}.} \bibinfo{year}{2023}\natexlab{}.
\newblock \showarticletitle{3d-llm: Injecting the 3d world into large language
  models}.
\newblock \bibinfo{journal}{\emph{Advances in Neural Information Processing
  Systems}}  \bibinfo{volume}{36} (\bibinfo{year}{2023}),
  \bibinfo{pages}{20482--20494}.
\newblock


\bibitem[Huang et~al\mbox{.}(2023)]%
        {huang2023chatscene}
\bibfield{author}{\bibinfo{person}{Haifeng Huang}, \bibinfo{person}{Yilun
  Chen}, \bibinfo{person}{Zehan Wang}, \bibinfo{person}{Rongjie Huang},
  \bibinfo{person}{Runsen Xu}, \bibinfo{person}{Tai Wang},
  \bibinfo{person}{Luping Liu}, \bibinfo{person}{Xize Cheng},
  \bibinfo{person}{Yang Zhao}, \bibinfo{person}{Jiangmiao Pang},
  {et~al\mbox{.}}} \bibinfo{year}{2023}\natexlab{}.
\newblock \showarticletitle{Chat-scene: Bridging 3d scene and large language
  models with object identifiers}.
\newblock \bibinfo{journal}{\emph{arXiv preprint arXiv:2312.08168}}
  (\bibinfo{year}{2023}).
\newblock


\bibitem[Hunyuan3D et~al\mbox{.}(2026)]%
        {hunyuan3d2026hy3dbench}
\bibfield{author}{\bibinfo{person}{Team Hunyuan3D}, \bibinfo{person}{Bowen
  Zhang}, \bibinfo{person}{Chunchao Guo}, \bibinfo{person}{Dongyuan Guo},
  \bibinfo{person}{Haolin Liu}, \bibinfo{person}{Hongyu Yan},
  \bibinfo{person}{Huiwen Shi}, \bibinfo{person}{Jiaao Yu},
  \bibinfo{person}{Jiachen Xu}, \bibinfo{person}{Jingwei Huang},
  {et~al\mbox{.}}} \bibinfo{year}{2026}\natexlab{}.
\newblock \showarticletitle{HY3D-Bench: Generation of 3D Assets}.
\newblock \bibinfo{journal}{\emph{arXiv preprint arXiv:2602.03907}}
  (\bibinfo{year}{2026}).
\newblock


\bibitem[Hurst et~al\mbox{.}(2024)]%
        {hurst2024gpt}
\bibfield{author}{\bibinfo{person}{Aaron Hurst}, \bibinfo{person}{Adam Lerer},
  \bibinfo{person}{Adam~P Goucher}, \bibinfo{person}{Adam Perelman},
  \bibinfo{person}{Aditya Ramesh}, \bibinfo{person}{Aidan Clark},
  \bibinfo{person}{AJ Ostrow}, \bibinfo{person}{Akila Welihinda},
  \bibinfo{person}{Alan Hayes}, \bibinfo{person}{Alec Radford},
  {et~al\mbox{.}}} \bibinfo{year}{2024}\natexlab{}.
\newblock \showarticletitle{Gpt-4o system card}.
\newblock \bibinfo{journal}{\emph{arXiv preprint arXiv:2410.21276}}
  (\bibinfo{year}{2024}).
\newblock


\bibitem[Jiang et~al\mbox{.}(2025)]%
        {rexomni}
\bibfield{author}{\bibinfo{person}{Qing Jiang}, \bibinfo{person}{Junan Huo},
  \bibinfo{person}{Xingyu Chen}, \bibinfo{person}{Yuda Xiong},
  \bibinfo{person}{Zhaoyang Zeng}, \bibinfo{person}{Yihao Chen},
  \bibinfo{person}{Tianhe Ren}, \bibinfo{person}{Junzhi Yu}, {and}
  \bibinfo{person}{Lei Zhang}.} \bibinfo{year}{2025}\natexlab{}.
\newblock \bibinfo{title}{Detect Anything via Next Point Prediction}.
\newblock
\showeprint[arxiv]{2510.12798}~[cs.CV]
\urldef\tempurl%
\url{https://arxiv.org/abs/2510.12798}
\showURL{%
\tempurl}


\bibitem[Jin et~al\mbox{.}(2026)]%
        {jin2026cosmo3d}
\bibfield{author}{\bibinfo{person}{Li Jin}, \bibinfo{person}{Weikai Chen},
  \bibinfo{person}{Yujie Wang}, \bibinfo{person}{Yingda Yin},
  \bibinfo{person}{Zeyu Hu}, \bibinfo{person}{Runze Zhang},
  \bibinfo{person}{Keyang Luo}, \bibinfo{person}{Shengju Qian},
  \bibinfo{person}{Xin Wang}, {and} \bibinfo{person}{Xueying Qin}.}
  \bibinfo{year}{2026}\natexlab{}.
\newblock \showarticletitle{CoSMo3D: Open-World Promptable 3D Semantic Part
  Segmentation through LLM-Guided Canonical Spatial Modeling}.
\newblock \bibinfo{journal}{\emph{arXiv preprint arXiv:2603.01205}}
  (\bibinfo{year}{2026}).
\newblock


\bibitem[Kirillov et~al\mbox{.}(2023)]%
        {sam1}
\bibfield{author}{\bibinfo{person}{Alexander Kirillov}, \bibinfo{person}{Eric
  Mintun}, \bibinfo{person}{Nikhila Ravi}, \bibinfo{person}{Hanzi Mao},
  \bibinfo{person}{Chloe Rolland}, \bibinfo{person}{Laura Gustafson},
  \bibinfo{person}{Tete Xiao}, \bibinfo{person}{Spencer Whitehead},
  \bibinfo{person}{Alexander~C Berg}, \bibinfo{person}{Wan-Yen Lo},
  {et~al\mbox{.}}} \bibinfo{year}{2023}\natexlab{}.
\newblock \showarticletitle{Segment anything}. In
  \bibinfo{booktitle}{\emph{Proceedings of the IEEE/CVF International
  Conference on Computer Vision}}. \bibinfo{pages}{4015--4026}.
\newblock


\bibitem[Kwon et~al\mbox{.}(2023)]%
        {vllm}
\bibfield{author}{\bibinfo{person}{Woosuk Kwon}, \bibinfo{person}{Zhuohan Li},
  \bibinfo{person}{Siyuan Zhuang}, \bibinfo{person}{Ying Sheng},
  \bibinfo{person}{Lianmin Zheng}, \bibinfo{person}{Cody~Hao Yu},
  \bibinfo{person}{Joseph~E. Gonzalez}, \bibinfo{person}{Hao Zhang}, {and}
  \bibinfo{person}{Ion Stoica}.} \bibinfo{year}{2023}\natexlab{}.
\newblock \showarticletitle{Efficient Memory Management for Large Language
  Model Serving with PagedAttention}. In \bibinfo{booktitle}{\emph{Proceedings
  of the ACM SIGOPS 29th Symposium on Operating Systems Principles}}.
\newblock


\bibitem[Lai et~al\mbox{.}(2024)]%
        {lai2024lisa}
\bibfield{author}{\bibinfo{person}{Xin Lai}, \bibinfo{person}{Zhuotao Tian},
  \bibinfo{person}{Yukang Chen}, \bibinfo{person}{Yanwei Li},
  \bibinfo{person}{Yuhui Yuan}, \bibinfo{person}{Shu Liu}, {and}
  \bibinfo{person}{Jiaya Jia}.} \bibinfo{year}{2024}\natexlab{}.
\newblock \showarticletitle{Lisa: Reasoning segmentation via large language
  model}. In \bibinfo{booktitle}{\emph{Proceedings of the IEEE/CVF conference
  on computer vision and pattern recognition}}. \bibinfo{pages}{9579--9589}.
\newblock


\bibitem[Lai et~al\mbox{.}(2025)]%
        {hunyuan3d}
\bibfield{author}{\bibinfo{person}{Zeqiang Lai}, \bibinfo{person}{Yunfei Zhao},
  \bibinfo{person}{Haolin Liu}, \bibinfo{person}{Zibo Zhao},
  \bibinfo{person}{Qingxiang Lin}, \bibinfo{person}{Huiwen Shi},
  \bibinfo{person}{Xianghui Yang}, \bibinfo{person}{Mingxin Yang},
  \bibinfo{person}{Shuhui Yang}, \bibinfo{person}{Yifei Feng}, {et~al\mbox{.}}}
  \bibinfo{year}{2025}\natexlab{}.
\newblock \showarticletitle{Hunyuan3D 2.5: Towards High-Fidelity 3D Assets
  Generation with Ultimate Details}.
\newblock \bibinfo{journal}{\emph{arXiv preprint arXiv:2506.16504}}
  (\bibinfo{year}{2025}).
\newblock


\bibitem[Lang et~al\mbox{.}(2024)]%
        {lang2024iseg}
\bibfield{author}{\bibinfo{person}{Itai Lang}, \bibinfo{person}{Fei Xu},
  \bibinfo{person}{Dale Decatur}, \bibinfo{person}{Sudarshan Babu}, {and}
  \bibinfo{person}{Rana Hanocka}.} \bibinfo{year}{2024}\natexlab{}.
\newblock \showarticletitle{iseg: Interactive 3d segmentation via interactive
  attention}. In \bibinfo{booktitle}{\emph{SIGGRAPH Asia 2024 Conference
  Papers}}. \bibinfo{pages}{1--11}.
\newblock


\bibitem[Li et~al\mbox{.}(2026)]%
        {li2026segvigen}
\bibfield{author}{\bibinfo{person}{Lin Li}, \bibinfo{person}{Haoran Feng},
  \bibinfo{person}{Zehuan Huang}, \bibinfo{person}{Haohua Chen},
  \bibinfo{person}{Wenbo Nie}, \bibinfo{person}{Shaohua Hou},
  \bibinfo{person}{Keqing Fan}, \bibinfo{person}{Pan Hu},
  \bibinfo{person}{Sheng Wang}, \bibinfo{person}{Buyu Li}, {and}
  \bibinfo{person}{Lu Sheng}.} \bibinfo{year}{2026}\natexlab{}.
\newblock \showarticletitle{SegviGen: Repurposing 3D Generative Model for Part
  Segmentation}.
\newblock \bibinfo{journal}{\emph{ACM Trans. Graph.}} \bibinfo{volume}{45},
  \bibinfo{number}{4}, Article \bibinfo{articleno}{68} (\bibinfo{date}{July}
  \bibinfo{year}{2026}), \bibinfo{numpages}{12}~pages.
\newblock
\showISSN{0730-0301}
\href{https://doi.org/10.1145/3811399}{doi:\nolinkurl{10.1145/3811399}}


\bibitem[Li et~al\mbox{.}(2022)]%
        {li2022grounded}
\bibfield{author}{\bibinfo{person}{Liunian~Harold Li},
  \bibinfo{person}{Pengchuan Zhang}, \bibinfo{person}{Haotian Zhang},
  \bibinfo{person}{Jianwei Yang}, \bibinfo{person}{Chunyuan Li},
  \bibinfo{person}{Yiwu Zhong}, \bibinfo{person}{Lijuan Wang},
  \bibinfo{person}{Lu Yuan}, \bibinfo{person}{Lei Zhang},
  \bibinfo{person}{Jenq-Neng Hwang}, {et~al\mbox{.}}}
  \bibinfo{year}{2022}\natexlab{}.
\newblock \showarticletitle{Grounded language-image pre-training}. In
  \bibinfo{booktitle}{\emph{Proceedings of the IEEE/CVF Conference on Computer
  Vision and Pattern Recognition}}. \bibinfo{pages}{10965--10975}.
\newblock


\bibitem[Liu et~al\mbox{.}(2026)]%
        {liu2026quadgpt}
\bibfield{author}{\bibinfo{person}{Jian Liu}, \bibinfo{person}{Chunshi Wang},
  \bibinfo{person}{Song Guo}, \bibinfo{person}{Haohan Weng},
  \bibinfo{person}{Zhen Zhou}, \bibinfo{person}{Zhiqi Li},
  \bibinfo{person}{Jiaao Yu}, \bibinfo{person}{Yiling Zhu},
  \bibinfo{person}{Jing Xu}, \bibinfo{person}{Biwen Lei}, \bibinfo{person}{Zhuo
  Chen}, {and} \bibinfo{person}{Chunchao Guo}.}
  \bibinfo{year}{2026}\natexlab{}.
\newblock \showarticletitle{Quad{GPT}: Native Quadrilateral Mesh Generation
  with Autoregressive Models}. In \bibinfo{booktitle}{\emph{The Fourteenth
  International Conference on Learning Representations}}.
\newblock
\urldef\tempurl%
\url{https://openreview.net/forum?id=oRmo4p1KEE}
\showURL{%
\tempurl}


\bibitem[Liu et~al\mbox{.}(2022)]%
        {liu2022akb}
\bibfield{author}{\bibinfo{person}{Liu Liu}, \bibinfo{person}{Wenqiang Xu},
  \bibinfo{person}{Haoyuan Fu}, \bibinfo{person}{Sucheng Qian},
  \bibinfo{person}{Qiaojun Yu}, \bibinfo{person}{Yang Han}, {and}
  \bibinfo{person}{Cewu Lu}.} \bibinfo{year}{2022}\natexlab{}.
\newblock \showarticletitle{Akb-48: A real-world articulated object knowledge
  base}. In \bibinfo{booktitle}{\emph{Proceedings of the IEEE/CVF Conference on
  Computer Vision and Pattern Recognition}}. \bibinfo{pages}{14789--14798}.
\newblock


\bibitem[Liu et~al\mbox{.}(2025)]%
        {liu2025partfield}
\bibfield{author}{\bibinfo{person}{Minghua Liu},
  \bibinfo{person}{Mikaela~Angelina Uy}, \bibinfo{person}{Donglai Xiang},
  \bibinfo{person}{Hao Su}, \bibinfo{person}{Sanja Fidler},
  \bibinfo{person}{Nicholas Sharp}, {and} \bibinfo{person}{Jun Gao}.}
  \bibinfo{year}{2025}\natexlab{}.
\newblock \showarticletitle{PartField: Learning 3D Feature Fields for Part
  Segmentation and Beyond}. In \bibinfo{booktitle}{\emph{Proceedings of the
  IEEE/CVF International Conference on Computer Vision}}.
  \bibinfo{pages}{9704--9715}.
\newblock


\bibitem[Liu et~al\mbox{.}(2023)]%
        {liu2023partslip}
\bibfield{author}{\bibinfo{person}{Minghua Liu}, \bibinfo{person}{Yinhao Zhu},
  \bibinfo{person}{Hong Cai}, \bibinfo{person}{Shizhong Han},
  \bibinfo{person}{Zhan Ling}, \bibinfo{person}{Fatih Porikli}, {and}
  \bibinfo{person}{Hao Su}.} \bibinfo{year}{2023}\natexlab{}.
\newblock \showarticletitle{Partslip: Low-shot part segmentation for 3d point
  clouds via pretrained image-language models}. In
  \bibinfo{booktitle}{\emph{Proceedings of the IEEE/CVF Conference on Computer
  Vision and Pattern Recognition}}. \bibinfo{pages}{21736--21746}.
\newblock


\bibitem[Ma et~al\mbox{.}(2025)]%
        {p3sam}
\bibfield{author}{\bibinfo{person}{Changfeng Ma}, \bibinfo{person}{Yang Li},
  \bibinfo{person}{Xinhao Yan}, \bibinfo{person}{Jiachen Xu},
  \bibinfo{person}{Yunhan Yang}, \bibinfo{person}{Chunshi Wang},
  \bibinfo{person}{Zibo Zhao}, \bibinfo{person}{Yanwen Guo},
  \bibinfo{person}{Zhuo Chen}, {and} \bibinfo{person}{Chunchao Guo}.}
  \bibinfo{year}{2025}\natexlab{}.
\newblock \showarticletitle{P3-sam: Native 3d part segmentation}.
\newblock \bibinfo{journal}{\emph{arXiv preprint arXiv:2509.06784}}
  (\bibinfo{year}{2025}).
\newblock


\bibitem[Ma et~al\mbox{.}(2024)]%
        {ma2024find}
\bibfield{author}{\bibinfo{person}{Ziqi Ma}, \bibinfo{person}{Yisong Yue},
  {and} \bibinfo{person}{Georgia Gkioxari}.} \bibinfo{year}{2024}\natexlab{}.
\newblock \showarticletitle{Find any part in 3d}.
\newblock \bibinfo{journal}{\emph{arXiv preprint arXiv:2411.13550}}
  (\bibinfo{year}{2024}).
\newblock


\bibitem[Mao et~al\mbox{.}(2026)]%
        {mao2025spatiallm}
\bibfield{author}{\bibinfo{person}{Yongsen Mao}, \bibinfo{person}{Junhao
  Zhong}, \bibinfo{person}{Chuan Fang}, \bibinfo{person}{Jia Zheng},
  \bibinfo{person}{Rui Tang}, \bibinfo{person}{Hao Zhu}, \bibinfo{person}{Ping
  Tan}, {and} \bibinfo{person}{Zihan Zhou}.} \bibinfo{year}{2026}\natexlab{}.
\newblock \showarticletitle{Spatiallm: Training large language models for
  structured indoor modeling}.
\newblock \bibinfo{journal}{\emph{Advances in Neural Information Processing
  Systems}}  \bibinfo{volume}{38} (\bibinfo{year}{2026}),
  \bibinfo{pages}{45165--45195}.
\newblock


\bibitem[Mo et~al\mbox{.}(2019)]%
        {mo2019partnet}
\bibfield{author}{\bibinfo{person}{Kaichun Mo}, \bibinfo{person}{Shilin Zhu},
  \bibinfo{person}{Angel~X Chang}, \bibinfo{person}{Li Yi},
  \bibinfo{person}{Subarna Tripathi}, \bibinfo{person}{Leonidas~J Guibas},
  {and} \bibinfo{person}{Hao Su}.} \bibinfo{year}{2019}\natexlab{}.
\newblock \showarticletitle{Partnet: A large-scale benchmark for fine-grained
  and hierarchical part-level 3d object understanding}. In
  \bibinfo{booktitle}{\emph{Proceedings of the IEEE/CVF Conference on Computer
  Vision and Pattern Recognition}}. \bibinfo{pages}{909--918}.
\newblock


\bibitem[Oquab et~al\mbox{.}(2024)]%
        {oquab2024dinov2}
\bibfield{author}{\bibinfo{person}{Maxime Oquab}, \bibinfo{person}{Timoth{\'e}e
  Darcet}, \bibinfo{person}{Th{\'e}o Moutakanni}, \bibinfo{person}{Huy~V. Vo},
  \bibinfo{person}{Marc Szafraniec}, \bibinfo{person}{Vasil Khalidov},
  \bibinfo{person}{Pierre Fernandez}, \bibinfo{person}{Daniel HAZIZA},
  \bibinfo{person}{Francisco Massa}, \bibinfo{person}{Alaaeldin El-Nouby},
  \bibinfo{person}{Mido Assran}, \bibinfo{person}{Nicolas Ballas},
  \bibinfo{person}{Wojciech Galuba}, \bibinfo{person}{Russell Howes},
  \bibinfo{person}{Po-Yao Huang}, \bibinfo{person}{Shang-Wen Li},
  \bibinfo{person}{Ishan Misra}, \bibinfo{person}{Michael Rabbat},
  \bibinfo{person}{Vasu Sharma}, \bibinfo{person}{Gabriel Synnaeve},
  \bibinfo{person}{Hu Xu}, \bibinfo{person}{Herve Jegou},
  \bibinfo{person}{Julien Mairal}, \bibinfo{person}{Patrick Labatut},
  \bibinfo{person}{Armand Joulin}, {and} \bibinfo{person}{Piotr Bojanowski}.}
  \bibinfo{year}{2024}\natexlab{}.
\newblock \showarticletitle{{DINO}v2: Learning Robust Visual Features without
  Supervision}.
\newblock \bibinfo{journal}{\emph{Transactions on Machine Learning Research}}
  (\bibinfo{year}{2024}).
\newblock
\showISSN{2835-8856}
\urldef\tempurl%
\url{https://openreview.net/forum?id=a68SUt6zFt}
\showURL{%
\tempurl}
\newblock
\shownote{Featured Certification}.


\bibitem[Paul et~al\mbox{.}(2025)]%
        {paul2025name}
\bibfield{author}{\bibinfo{person}{Soumava Paul}, \bibinfo{person}{Prakhar
  Kaushik}, \bibinfo{person}{Ankit Vaidya}, \bibinfo{person}{Anand Bhattad},
  {and} \bibinfo{person}{Alan Yuille}.} \bibinfo{year}{2025}\natexlab{}.
\newblock \showarticletitle{Name That Part: 3D Part Segmentation and Naming}.
\newblock \bibinfo{journal}{\emph{arXiv preprint arXiv:2512.18003}}
  (\bibinfo{year}{2025}).
\newblock


\bibitem[Peng et~al\mbox{.}(2024)]%
        {peng2023kosmos}
\bibfield{author}{\bibinfo{person}{Zhiliang Peng}, \bibinfo{person}{Wenhui
  Wang}, \bibinfo{person}{Li Dong}, \bibinfo{person}{Yaru Hao},
  \bibinfo{person}{Shaohan Huang}, \bibinfo{person}{Shuming Ma},
  \bibinfo{person}{Qixiang Ye}, {and} \bibinfo{person}{Furu Wei}.}
  \bibinfo{year}{2024}\natexlab{}.
\newblock \showarticletitle{Grounding multimodal large language models to the
  world}. In \bibinfo{booktitle}{\emph{International Conference on Learning
  Representations}}, Vol.~\bibinfo{volume}{2024}.
  \bibinfo{pages}{51575--51598}.
\newblock


\bibitem[Qi et~al\mbox{.}(2017)]%
        {qi2017pointnet++}
\bibfield{author}{\bibinfo{person}{Charles~Ruizhongtai Qi}, \bibinfo{person}{Li
  Yi}, \bibinfo{person}{Hao Su}, {and} \bibinfo{person}{Leonidas~J Guibas}.}
  \bibinfo{year}{2017}\natexlab{}.
\newblock \showarticletitle{Pointnet++: Deep hierarchical feature learning on
  point sets in a metric space}.
\newblock \bibinfo{journal}{\emph{Advances in Neural Information Processing
  Systems}}  \bibinfo{volume}{30} (\bibinfo{year}{2017}).
\newblock


\bibitem[Radford et~al\mbox{.}(2021)]%
        {clip}
\bibfield{author}{\bibinfo{person}{Alec Radford}, \bibinfo{person}{Jong~Wook
  Kim}, \bibinfo{person}{Chris Hallacy}, \bibinfo{person}{Aditya Ramesh},
  \bibinfo{person}{Gabriel Goh}, \bibinfo{person}{Sandhini Agarwal},
  \bibinfo{person}{Girish Sastry}, \bibinfo{person}{Amanda Askell},
  \bibinfo{person}{Pamela Mishkin}, \bibinfo{person}{Jack Clark},
  {et~al\mbox{.}}} \bibinfo{year}{2021}\natexlab{}.
\newblock \showarticletitle{Learning transferable visual models from natural
  language supervision}. In \bibinfo{booktitle}{\emph{International Conference
  on Machine Learning}}. \bibinfo{pages}{8748--8763}.
\newblock


\bibitem[Rasheed et~al\mbox{.}(2024)]%
        {rasheed2024glamm}
\bibfield{author}{\bibinfo{person}{Hanoona Rasheed}, \bibinfo{person}{Muhammad
  Maaz}, \bibinfo{person}{Sahal Shaji}, \bibinfo{person}{Abdelrahman Shaker},
  \bibinfo{person}{Salman Khan}, \bibinfo{person}{Hisham Cholakkal},
  \bibinfo{person}{Rao~M Anwer}, \bibinfo{person}{Eric Xing},
  \bibinfo{person}{Ming-Hsuan Yang}, {and} \bibinfo{person}{Fahad~S Khan}.}
  \bibinfo{year}{2024}\natexlab{}.
\newblock \showarticletitle{Glamm: Pixel grounding large multimodal model}. In
  \bibinfo{booktitle}{\emph{Proceedings of the IEEE/CVF Conference on Computer
  Vision and Pattern Recognition}}. \bibinfo{pages}{13009--13018}.
\newblock


\bibitem[Ren et~al\mbox{.}(2024)]%
        {ren2024pixellm}
\bibfield{author}{\bibinfo{person}{Zhongwei Ren}, \bibinfo{person}{Zhicheng
  Huang}, \bibinfo{person}{Yunchao Wei}, \bibinfo{person}{Yao Zhao},
  \bibinfo{person}{Dongmei Fu}, \bibinfo{person}{Jiashi Feng}, {and}
  \bibinfo{person}{Xiaojie Jin}.} \bibinfo{year}{2024}\natexlab{}.
\newblock \showarticletitle{Pixellm: Pixel reasoning with large multimodal
  model}. In \bibinfo{booktitle}{\emph{Proceedings of the IEEE/CVF Conference
  on Computer Vision and Pattern Recognition}}. \bibinfo{pages}{26374--26383}.
\newblock


\bibitem[Su et~al\mbox{.}(2025)]%
        {s2am3d}
\bibfield{author}{\bibinfo{person}{Han Su}, \bibinfo{person}{Tianyu Huang},
  \bibinfo{person}{Zichen Wan}, \bibinfo{person}{Xiaohe Wu}, {and}
  \bibinfo{person}{Wangmeng Zuo}.} \bibinfo{year}{2025}\natexlab{}.
\newblock \showarticletitle{S2AM3D: Scale-controllable Part Segmentation of 3D
  Point Cloud}.
\newblock \bibinfo{journal}{\emph{arXiv preprint arXiv:2512.00995}}
  (\bibinfo{year}{2025}).
\newblock


\bibitem[Tang et~al\mbox{.}(2024)]%
        {tang2024segment}
\bibfield{author}{\bibinfo{person}{George Tang}, \bibinfo{person}{William
  Zhao}, \bibinfo{person}{Logan Ford}, \bibinfo{person}{David Benhaim}, {and}
  \bibinfo{person}{Paul Zhang}.} \bibinfo{year}{2024}\natexlab{}.
\newblock \showarticletitle{Segment any mesh}.
\newblock \bibinfo{journal}{\emph{arXiv preprint arXiv:2408.13679}}
  (\bibinfo{year}{2024}).
\newblock


\bibitem[Thomas et~al\mbox{.}(2019)]%
        {thomas2019kpconv}
\bibfield{author}{\bibinfo{person}{Hugues Thomas}, \bibinfo{person}{Charles~R
  Qi}, \bibinfo{person}{Jean-Emmanuel Deschaud}, \bibinfo{person}{Beatriz
  Marcotegui}, \bibinfo{person}{Fran{\c{c}}ois Goulette}, {and}
  \bibinfo{person}{Leonidas~J Guibas}.} \bibinfo{year}{2019}\natexlab{}.
\newblock \showarticletitle{Kpconv: Flexible and deformable convolution for
  point clouds}. In \bibinfo{booktitle}{\emph{Proceedings of the IEEE/CVF
  International Conference on Computer Vision}}. \bibinfo{pages}{6411--6420}.
\newblock


\bibitem[Wang et~al\mbox{.}(2026b)]%
        {wang2025partxmllm}
\bibfield{author}{\bibinfo{person}{Chunshi Wang}, \bibinfo{person}{Junliang
  Ye}, \bibinfo{person}{Yunhan Yang}, \bibinfo{person}{YANG LI},
  \bibinfo{person}{Zizhuo Lin}, \bibinfo{person}{Jun Zhu},
  \bibinfo{person}{Zhuo Chen}, \bibinfo{person}{Yawei Luo}, {and}
  \bibinfo{person}{Chunchao Guo}.} \bibinfo{year}{2026}\natexlab{b}.
\newblock \showarticletitle{Part-X-{MLLM}: Part-aware 3D Multimodal Large
  Language Model}. In \bibinfo{booktitle}{\emph{The Fourteenth International
  Conference on Learning Representations}}.
\newblock
\urldef\tempurl%
\url{https://openreview.net/forum?id=WffiETiSeU}
\showURL{%
\tempurl}


\bibitem[Wang et~al\mbox{.}(2026a)]%
        {wang2026partnext}
\bibfield{author}{\bibinfo{person}{Penghao Wang}, \bibinfo{person}{Yiyang He},
  \bibinfo{person}{Xin Lv}, \bibinfo{person}{Yukai Zhou}, \bibinfo{person}{Lan
  Xu}, \bibinfo{person}{Jingyi Yu}, {and} \bibinfo{person}{Jiayuan Gu}.}
  \bibinfo{year}{2026}\natexlab{a}.
\newblock \showarticletitle{PartNeXt: A Next-Generation Dataset for
  Fine-Grained and Hierarchical 3D Part Understanding}. In
  \bibinfo{booktitle}{\emph{The Thirty-ninth Annual Conference on Neural
  Information Processing Systems Datasets and Benchmarks Track}}.
\newblock
\urldef\tempurl%
\url{https://openreview.net/forum?id=J0PmRFMZXm}
\showURL{%
\tempurl}


\bibitem[Wang et~al\mbox{.}(2019)]%
        {dgcnn}
\bibfield{author}{\bibinfo{person}{Yue Wang}, \bibinfo{person}{Yongbin Sun},
  \bibinfo{person}{Ziwei Liu}, \bibinfo{person}{Sanjay~E. Sarma},
  \bibinfo{person}{Michael~M. Bronstein}, {and} \bibinfo{person}{Justin~M.
  Solomon}.} \bibinfo{year}{2019}\natexlab{}.
\newblock \showarticletitle{Dynamic Graph CNN for Learning on Point Clouds}.
\newblock \bibinfo{journal}{\emph{ACM Trans. Graph.}} \bibinfo{volume}{38},
  \bibinfo{number}{5}, Article \bibinfo{articleno}{146} (\bibinfo{date}{Oct.}
  \bibinfo{year}{2019}), \bibinfo{numpages}{12}~pages.
\newblock
\showISSN{0730-0301}
\href{https://doi.org/10.1145/3326362}{doi:\nolinkurl{10.1145/3326362}}


\bibitem[Wu et~al\mbox{.}(2024)]%
        {ptv3}
\bibfield{author}{\bibinfo{person}{Xiaoyang Wu}, \bibinfo{person}{Li Jiang},
  \bibinfo{person}{Peng-Shuai Wang}, \bibinfo{person}{Zhijian Liu},
  \bibinfo{person}{Xihui Liu}, \bibinfo{person}{Yu Qiao},
  \bibinfo{person}{Wanli Ouyang}, \bibinfo{person}{Tong He}, {and}
  \bibinfo{person}{Hengshuang Zhao}.} \bibinfo{year}{2024}\natexlab{}.
\newblock \showarticletitle{Point transformer v3: Simpler faster stronger}. In
  \bibinfo{booktitle}{\emph{Proceedings of the IEEE/CVF conference on computer
  vision and pattern recognition}}. \bibinfo{pages}{4840--4851}.
\newblock


\bibitem[Xiang et~al\mbox{.}(2025)]%
        {trellis2}
\bibfield{author}{\bibinfo{person}{Jianfeng Xiang}, \bibinfo{person}{Xiaoxue
  Chen}, \bibinfo{person}{Sicheng Xu}, \bibinfo{person}{Ruicheng Wang},
  \bibinfo{person}{Zelong Lv}, \bibinfo{person}{Yu Deng},
  \bibinfo{person}{Hongyuan Zhu}, \bibinfo{person}{Yue Dong},
  \bibinfo{person}{Hao Zhao}, \bibinfo{person}{Nicholas~Jing Yuan},
  {et~al\mbox{.}}} \bibinfo{year}{2025}\natexlab{}.
\newblock \showarticletitle{Native and compact structured latents for 3d
  generation}.
\newblock \bibinfo{journal}{\emph{arXiv preprint arXiv:2512.14692}}
  (\bibinfo{year}{2025}).
\newblock


\bibitem[Xue et~al\mbox{.}(2025)]%
        {xue2025zerops}
\bibfield{author}{\bibinfo{person}{Yuheng Xue}, \bibinfo{person}{Nenglun Chen},
  \bibinfo{person}{Jun Liu}, {and} \bibinfo{person}{Wenyun Sun}.}
  \bibinfo{year}{2025}\natexlab{}.
\newblock \showarticletitle{Zerops: High-quality cross-modal knowledge transfer
  for zero-shot 3d part segmentation}. In \bibinfo{booktitle}{\emph{2025
  International Conference on 3D Vision}}. \bibinfo{pages}{1328--1339}.
\newblock


\bibitem[Yang et~al\mbox{.}(2024)]%
        {yang2024sampart3d}
\bibfield{author}{\bibinfo{person}{Yunhan Yang}, \bibinfo{person}{Yukun Huang},
  \bibinfo{person}{Yuan-Chen Guo}, \bibinfo{person}{Liangjun Lu},
  \bibinfo{person}{Xiaoyang Wu}, \bibinfo{person}{Edmund~Y Lam},
  \bibinfo{person}{Yan-Pei Cao}, {and} \bibinfo{person}{Xihui Liu}.}
  \bibinfo{year}{2024}\natexlab{}.
\newblock \showarticletitle{Sampart3d: Segment any part in 3d objects}.
\newblock \bibinfo{journal}{\emph{arXiv preprint arXiv:2411.07184}}
  (\bibinfo{year}{2024}).
\newblock


\bibitem[Yi et~al\mbox{.}(2016)]%
        {shapenetpart}
\bibfield{author}{\bibinfo{person}{Li Yi}, \bibinfo{person}{Vladimir~G. Kim},
  \bibinfo{person}{Duygu Ceylan}, \bibinfo{person}{I-Chao Shen},
  \bibinfo{person}{Mengyan Yan}, \bibinfo{person}{Hao Su},
  \bibinfo{person}{Cewu Lu}, \bibinfo{person}{Qixing Huang},
  \bibinfo{person}{Alla Sheffer}, {and} \bibinfo{person}{Leonidas Guibas}.}
  \bibinfo{year}{2016}\natexlab{}.
\newblock \showarticletitle{A scalable active framework for region annotation
  in 3D shape collections}.
\newblock \bibinfo{journal}{\emph{ACM Trans. Graph.}} \bibinfo{volume}{35},
  \bibinfo{number}{6}, Article \bibinfo{articleno}{210} (\bibinfo{date}{Dec.}
  \bibinfo{year}{2016}), \bibinfo{numpages}{12}~pages.
\newblock
\showISSN{0730-0301}
\href{https://doi.org/10.1145/2980179.2980238}{doi:\nolinkurl{10.1145/2980179.2980238}}


\bibitem[You et~al\mbox{.}(2024)]%
        {you2023ferret}
\bibfield{author}{\bibinfo{person}{Haoxuan You}, \bibinfo{person}{Haotian
  Zhang}, \bibinfo{person}{Zhe Gan}, \bibinfo{person}{Xianzhi Du},
  \bibinfo{person}{Bowen Zhang}, \bibinfo{person}{Zirui Wang},
  \bibinfo{person}{Liangliang Cao}, \bibinfo{person}{Shih-Fu Chang}, {and}
  \bibinfo{person}{Yinfei Yang}.} \bibinfo{year}{2024}\natexlab{}.
\newblock \showarticletitle{Ferret: Refer and ground anything anywhere at any
  granularity}. In \bibinfo{booktitle}{\emph{International Conference on
  Learning Representations}}, Vol.~\bibinfo{volume}{2024}.
  \bibinfo{pages}{57153--57180}.
\newblock


\bibitem[Zhang et~al\mbox{.}(2026)]%
        {zhang2026utonia}
\bibfield{author}{\bibinfo{person}{Yujia Zhang}, \bibinfo{person}{Xiaoyang Wu},
  \bibinfo{person}{Yunhan Yang}, \bibinfo{person}{Xianzhe Fan},
  \bibinfo{person}{Han Li}, \bibinfo{person}{Yuechen Zhang},
  \bibinfo{person}{Zehao Huang}, \bibinfo{person}{Naiyan Wang}, {and}
  \bibinfo{person}{Hengshuang Zhao}.} \bibinfo{year}{2026}\natexlab{}.
\newblock \showarticletitle{Utonia: Toward One Encoder for All Point Clouds}.
\newblock \bibinfo{journal}{\emph{arXiv preprint arXiv:2603.03283}}
  (\bibinfo{year}{2026}).
\newblock


\bibitem[Zhao et~al\mbox{.}(2021)]%
        {zhao2021point}
\bibfield{author}{\bibinfo{person}{Hengshuang Zhao}, \bibinfo{person}{Li
  Jiang}, \bibinfo{person}{Jiaya Jia}, \bibinfo{person}{Philip~HS Torr}, {and}
  \bibinfo{person}{Vladlen Koltun}.} \bibinfo{year}{2021}\natexlab{}.
\newblock \showarticletitle{Point transformer}. In
  \bibinfo{booktitle}{\emph{Proceedings of the IEEE/CVF International
  Conference on Computer Vision}}. \bibinfo{pages}{16259--16268}.
\newblock


\bibitem[Zhao et~al\mbox{.}(2025)]%
        {zhao2025deepmesh}
\bibfield{author}{\bibinfo{person}{Ruowen Zhao}, \bibinfo{person}{Junliang Ye},
  \bibinfo{person}{Zhengyi Wang}, \bibinfo{person}{Guangce Liu},
  \bibinfo{person}{Yiwen Chen}, \bibinfo{person}{Yikai Wang}, {and}
  \bibinfo{person}{Jun Zhu}.} \bibinfo{year}{2025}\natexlab{}.
\newblock \showarticletitle{Deepmesh: Auto-regressive artist-mesh creation with
  reinforcement learning}. In \bibinfo{booktitle}{\emph{Proceedings of the
  IEEE/CVF International Conference on Computer Vision}}.
  \bibinfo{pages}{10612--10623}.
\newblock


\bibitem[Zhong et~al\mbox{.}(2024)]%
        {zhong2024meshsegmenter}
\bibfield{author}{\bibinfo{person}{Ziming Zhong}, \bibinfo{person}{Yanyu Xu},
  \bibinfo{person}{Jing Li}, \bibinfo{person}{Jiale Xu},
  \bibinfo{person}{Zhengxin Li}, \bibinfo{person}{Chaohui Yu}, {and}
  \bibinfo{person}{Shenghua Gao}.} \bibinfo{year}{2024}\natexlab{}.
\newblock \showarticletitle{Meshsegmenter: Zero-shot mesh semantic segmentation
  via texture synthesis}. In \bibinfo{booktitle}{\emph{European Conference on
  Computer Vision}}. \bibinfo{pages}{182--199}.
\newblock


\bibitem[Zhou et~al\mbox{.}(2025)]%
        {zhou2025pointsam}
\bibfield{author}{\bibinfo{person}{Yuchen Zhou}, \bibinfo{person}{Jiayuan Gu},
  \bibinfo{person}{Tung~Yen Chiang}, \bibinfo{person}{Fanbo Xiang}, {and}
  \bibinfo{person}{Hao Su}.} \bibinfo{year}{2025}\natexlab{}.
\newblock \showarticletitle{Point-{SAM}: Promptable 3D Segmentation Model for
  Point Clouds}. In \bibinfo{booktitle}{\emph{The Thirteenth International
  Conference on Learning Representations}}.
\newblock


\bibitem[Zhou et~al\mbox{.}(2023)]%
        {zhou2023partslip++}
\bibfield{author}{\bibinfo{person}{Yuchen Zhou}, \bibinfo{person}{Jiayuan Gu},
  \bibinfo{person}{Xuanlin Li}, \bibinfo{person}{Minghua Liu},
  \bibinfo{person}{Yunhao Fang}, {and} \bibinfo{person}{Hao Su}.}
  \bibinfo{year}{2023}\natexlab{}.
\newblock \showarticletitle{Partslip++: Enhancing low-shot 3d part segmentation
  via multi-view instance segmentation and maximum likelihood estimation}.
\newblock \bibinfo{journal}{\emph{arXiv preprint arXiv:2312.03015}}
  (\bibinfo{year}{2023}).
\newblock


\bibitem[Zhu et~al\mbox{.}(2025)]%
        {zhu2024llava3d}
\bibfield{author}{\bibinfo{person}{Chenming Zhu}, \bibinfo{person}{Tai Wang},
  \bibinfo{person}{Wenwei Zhang}, \bibinfo{person}{Jiangmiao Pang}, {and}
  \bibinfo{person}{Xihui Liu}.} \bibinfo{year}{2025}\natexlab{}.
\newblock \showarticletitle{Llava-3d: A simple yet effective pathway to
  empowering lmms with 3d capabilities}. In
  \bibinfo{booktitle}{\emph{Proceedings of the IEEE/CVF International
  Conference on Computer Vision}}. \bibinfo{pages}{4295--4305}.
\newblock


\bibitem[Zhu et~al\mbox{.}(2023)]%
        {zhu2023pointclip}
\bibfield{author}{\bibinfo{person}{Xiangyang Zhu}, \bibinfo{person}{Renrui
  Zhang}, \bibinfo{person}{Bowei He}, \bibinfo{person}{Ziyu Guo},
  \bibinfo{person}{Ziyao Zeng}, \bibinfo{person}{Zipeng Qin},
  \bibinfo{person}{Shanghang Zhang}, {and} \bibinfo{person}{Peng Gao}.}
  \bibinfo{year}{2023}\natexlab{}.
\newblock \showarticletitle{Pointclip v2: Prompting clip and gpt for powerful
  3d open-world learning}. In \bibinfo{booktitle}{\emph{Proceedings of the
  IEEE/CVF International Conference on Computer Vision}}.
  \bibinfo{pages}{2639--2650}.
\newblock


\bibitem[Zhu et~al\mbox{.}(2026)]%
        {zhu2026partsam}
\bibfield{author}{\bibinfo{person}{Zhe Zhu}, \bibinfo{person}{Le Wan},
  \bibinfo{person}{Rui Xu}, \bibinfo{person}{Yiheng Zhang},
  \bibinfo{person}{Honghua Chen}, \bibinfo{person}{Zhiyang Dou},
  \bibinfo{person}{Cheng Lin}, \bibinfo{person}{Yuan Liu}, {and}
  \bibinfo{person}{Mingqiang Wei}.} \bibinfo{year}{2026}\natexlab{}.
\newblock \showarticletitle{Part{SAM}: A Scalable Promptable Part Segmentation
  Model Trained on Native 3D Data}. In \bibinfo{booktitle}{\emph{The Fourteenth
  International Conference on Learning Representations}}.
\newblock
\urldef\tempurl%
\url{https://openreview.net/forum?id=y8sZUQPYXC}
\showURL{%
\tempurl}


\end{thebibliography}

\end{document}